\documentclass[preprint,review,12pt]{elsarticle}

\usepackage{graphics}
\usepackage{graphicx}
\usepackage{epsfig}

\usepackage{amssymb}

\usepackage{subfigure}
\usepackage[pdfborder=001]{hyperref}
\usepackage{booktabs}
\usepackage{caption}

\journal{Information Sciences}

\begin{document}

\begin{frontmatter}



\title{Multiple chaotic central pattern generators with learning for legged locomotion and malfunction compensation}


\author[au1]{Guanjiao~Ren}
\author[au1]{Weihai~Chen}
\author[au2]{Sakyasingha~Dasgupta}
\author[au2]{Christoph~Kolodziejski}
\author[au2]{Florentin~W\"org\"otter}
\author[au2,au3]{Poramate~Manoonpong
\corref{cor}}\ead{poma@mmmi.sdu.dk}
\cortext[cor]{Corresponding author. TEL: +45 6550 8698, FAX: +45 6615 7697}

\address[au1]{School of Automation Science and Electrical Engineering, Beijing University of Aeronautics and Astronautics, Beijing, 100191 China}
\address[au2]{Bernstein Center for Computational Neuroscience, \uppercase\expandafter{\romannumeral3} Physikalisches Institut - Biophysik, Georg-August-Universit\"at G\"ottingen, 37077 Germany}
\address[au3]{M{\ae}sk Mc-Kinney M{\o}ller Institute, University of Southern Denmark,
Campusvej 55, 5230 Odense M, Denmark}

\begin{abstract}
An originally chaotic system can be controlled into various periodic dynamics. When it is implemented into a legged robot's locomotion control as a central pattern generator (CPG), sophisticated gait patterns arise so that the robot can perform various walking behaviors.
However, such a single chaotic CPG controller has difficulties dealing with leg malfunction. Specifically, in the scenarios presented here, its movement permanently deviates from the desired trajectory.
To address this problem, we extend the single chaotic CPG to multiple CPGs with learning. The learning mechanism is based on a simulated annealing algorithm. In a normal situation, the CPGs synchronize and their dynamics are identical. With leg malfunction or disability, the CPGs lose synchronization leading to independent dynamics. In this case, the learning mechanism is applied to automatically adjust the remaining legs' oscillation frequencies so that the robot adapts its locomotion to deal with the malfunction. As a consequence, the trajectory produced by the multiple chaotic CPGs resembles the original trajectory far better than the one produced by only a single CPG. The performance of the system is evaluated first in a physical simulation of a quadruped as well as a hexapod robot and finally in a real six-legged walking machine called AMOS\uppercase\expandafter{\romannumeral2}. The experimental results presented here reveal that using multiple CPGs with learning is an effective approach for adaptive locomotion generation where, for instance, different body parts have to perform independent movements for malfunction compensation.
\end{abstract}


\end{frontmatter}


\section{Introduction}
\label{Intro}

Humans, mammals, insects, and other arthropods employ legs for movement. Common to all of them is that their walking pattern usually shows a high level of proficiency adapted to the different terrains of their natural habitat. Legged robots, on the other hand, have not yet achieved this level of performance.

Optimized biomechanics and (neural) control create these efficient and often very elegant walking patterns in animals and some robots have copied this strategy with varying levels of success. Many reports have demonstrated gait generations in animals which are achieved through oscillations originating from the spinal cord (vertebrate) or from different ganglions (invertebrate) \cite{delcomyn1999walking,cohen1982nature}. This is known as the concept of central pattern generators (CPGs) and has been applied to different types of legged robots, such as the bipedal robot designed by \citet{taga1991self} and \citet{aoi2006stability,shinya2012functional}, the quadruped robot {\it Tekken} by \citet{fukuoka2003adaptive,kimura2007adaptive}, the hexapod robots by \citet{arena2002multi,arena2004adaptive}, \citet{inagaki2003cpg,inagaki2006wave} and in our previous works \cite{manoonpong2008sensor,steingrube2010self,ren2012multiple}. Bio-inspired amphibious robots \cite{patel2006evolving,ijspeert2007swimming,crespi2013salamandra} and snake-like robots \cite{crespi2008online} also employed this kind of control strategy. Further details on CPG-based locomotion control have been reviewed in \citet{ijspeert2008central}.

CPG-based locomotion is directly inspired by the way animals control their movement. It has many advantages, such as distributed control, the ability to deal with redundancies, and fast control loops. It also allows modulation of locomotion by simple control signals \cite{ijspeert2008central,manoonpong2013neural}. When applied to robot control, we do not need to know the precise mechanical model of a robot. We can also easily integrate sensory information and adjust the control signal due to the simple structure of a CPG. Therefore, CPG-based control has already become an effective approach to perform legged locomotion in robots.

However, there are several problems yet to be solved. Although previous CPG-based algorithms can generate sophisticated gait patterns and deal with irregularities of the terrain to some extent \cite{manoonpong2013neural}, the problem of leg malfunction compensation in CPG-based control is still a challenging task. A troubling control problem can arise from the fact that the main controller usually contains CPGs which always control all legs with identical frequency \cite{hooper2000central}. If a robot suffers from leg failures, the other -- still functioning -- legs cannot immediately tune their oscillations appropriately. In contrast, insects can adjust the frequency of each leg individually \cite{bassler1998pattern,schilling2007hexapod}. If their legs are malfunctioning or disabled, they can still perform proper locomotion by changing the oscillation frequencies of the legs independently. Such a phenomenon also appears in mammals. For example, a cat can walk with the hind legs over a treadmill belt while the fore legs rest on a stationary platform \cite{rossignol2000locomotion} even after spinal cord injury. This indicates that the cat can independently adjust each leg's movement to achieve stable locomotion.

Traditional robotic methods for compensating leg malfunction are complicated \cite{yang1998fault,johnson2010disturbance}. They are mostly based on kinematics or dynamics models \cite{erden2007torque}.
Robots usually have to detect where a malfunction happens, then replan the gait pattern and choose another proper foot contact point. For different legs, the different foot trajectories are recalculated using inverse kinematics. Hence, all situations have to be considered and, as such, the procedure is computational intensive.

In contrast to the traditional control methods, we develop a CPG-based control strategy not only to generate multiple gaits but also to deal with leg malfunction.
Inspired by multiple oscillators found in the neural system of insects \cite{bassler1998pattern,daun2011inter,daun2011neuron}, we extend our previously proposed chaotic CPG controller \cite{steingrube2010self} to multiple CPGs, according to the number of legs of the robot.
The CPGs can be synchronized or desynchronized to produce uniform or non-uniform patterns, respectively.
If all CPGs are synchronized, the neural outputs are the same. If they are desynchronized, the neural outputs can oscillate at different frequencies.
Thus, if some joints are disabled, other legs can change their oscillation frequencies independently.
A simulated annealing (SA) based approach \cite{dowsland1993simulated,bertsimas1993simulated} is applied to our robots in order to learn a suitable combination of leg oscillation frequencies, allowing leg malfunction compensation to be achieved automatically.
Furthermore, the applications to a hexapod robot and a quadruped robot demonstrate the effectiveness of our proposed algorithm and its generalization properties.
To verify our algorithm in a real world application, our hexapod robot AMOS\uppercase\expandafter{\romannumeral2} is employed to evaluate the control strategy and learning. The proposed methods allow AMOS\uppercase\expandafter{\romannumeral2} to perform multiple gaits and to adapt its locomotion in case of disabled legs.
Therefore, the main contribution of this paper is a novel control strategy relying on multiple chaotic CPGs with an additional automatic learning mechanism for leg malfunction compensation.

This article is structured as follows.
Section~\ref{sect2} presents the overall control algorithm where the chaotic CPG is briefly introduced as a single oscillator. After which, we show how to design multiple CPGs and also state how the multiple CPGs synchronize and desynchronize with each other. Section~\ref{learn} introduces the learning algorithm (simulated annealing) and the principle of selecting a suitable combination of leg oscillation frequencies for malfunction compensation. Section~\ref{sect4} demonstrates the implementation of the proposed multiple CPGs and the learning strategy on simulated hexapod and quadruped robots. Section~\ref{sect5} introduces our real hexapod walking platform - AMOS\uppercase\expandafter{\romannumeral2}. The learning results obtained from simulation are applied to the robot and the effectiveness of the results is successfully verified. Section~\ref{sect6} discusses the results, and finally in Section~\ref{sect7} we present our conclusion.

\section{Multiple chaotic central pattern generators and synchronization mechanism}
\label{sect2}

Our multiple CPGs-based locomotion controller is derived from the chaotic CPG controller, introduced in  \cite{steingrube2010self}. First, we describe a single CPG oscillator and then show how it can be extended to multiple CPGs.
The synchronization and desynchronization mechanisms are also presented.
The multiple CPGs generate either different periodic patterns  independently, or they become synchronized and generate the same pattern.
Here, they will be synchronized for basic locomotion generation and desynchronized for malfunction compensation.

\subsection{Single chaotic CPG}

\begin{figure}
	\begin{center}
		\includegraphics[height=1.9in]{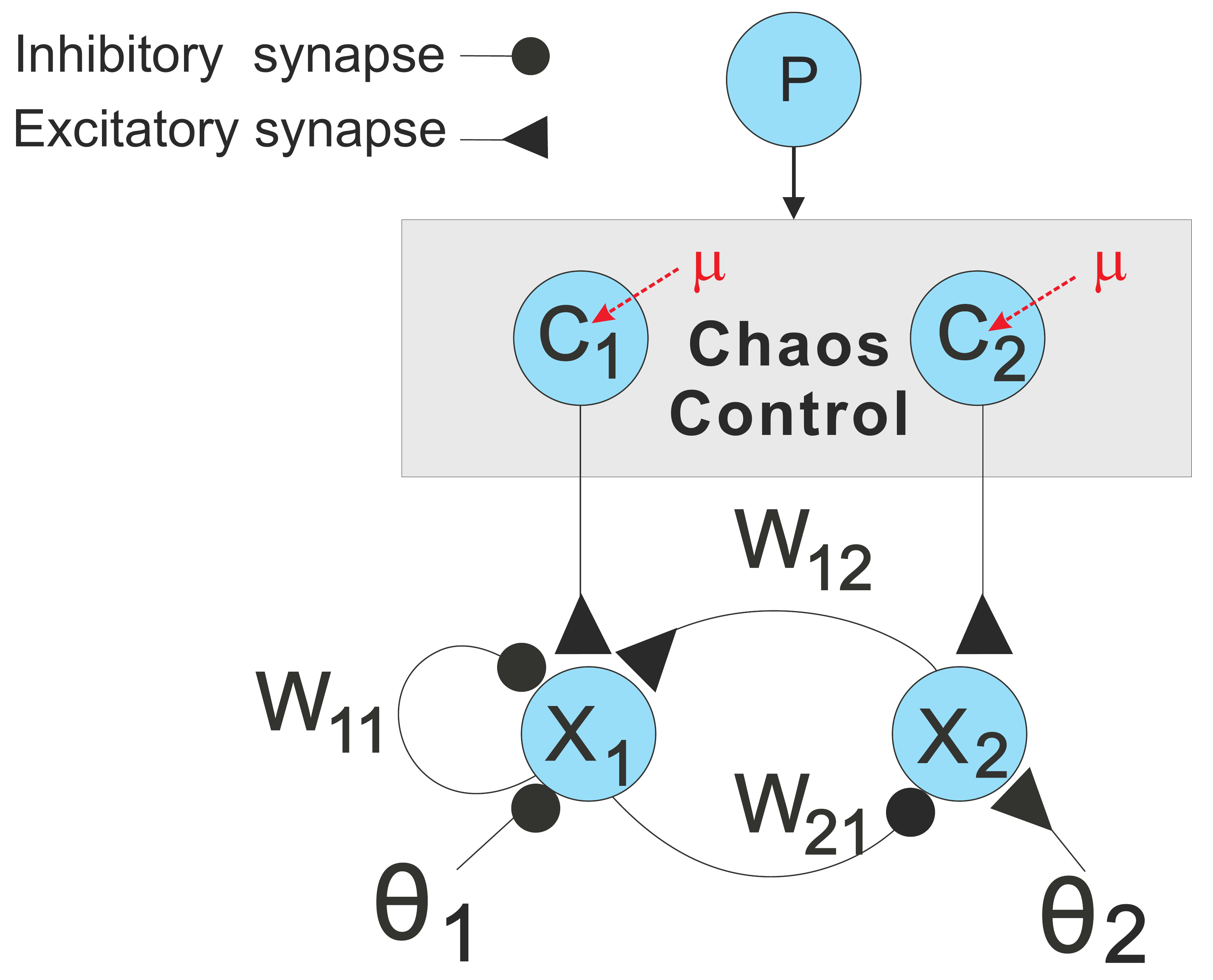}
	\hspace{0.4in}
	\caption{Single CPG with the chaos controller.  }
	\label{fig3}
	\end{center}
\end{figure}

The chaos control CPG unit is shown in Fig. \ref{fig3}.
In this figure, $x_1$ and $x_2$ indicate the neurons that generate the oscillation, while $c_1$ and $c_2$ are the control inputs depending only on the period $p$ with a control strength $\mu$. $w_{11}, w_{12}, w_{21}$ represent the synaptic weights and $\theta_1$ and $\theta_2$ indicate the biases.
Dynamics of the chaos control CPG can be exploited to generate complex patterns for legged robots, like chaotic leg motion and different walking patterns (multiple gaits).
To achieve different walking patterns, we simultaneously add inputs to the two neurons, i.e., the control signals $c_1$ and $c_2$.
These act as extra biases that depend on a single parameter {\it p} (the period of the output to be controlled).
The output of the neurons is detected every {\it p} steps and the chaos is controlled to a {\it p}-period orbit by adjusting the control input. The discrete time dynamics of the activity (output) states $x_i(t) \in [0,1]$ of the circuit satisfies:
\begin{equation}
	x_i(t+1) = \sigma(\theta_i + \sum_{j=1}^{2}w_{ij}x_j(t) + c_i^{(p)}(t)) \mbox{\space \space\space for } i\in\{1,2\} \label{eqa1}
\end{equation}
where $\sigma(x)=(1+\exp(-x))^{-1}$ is a sigmoid activation function with biases $\theta_i$ and $w_{ij}$ is the synaptic weight from neuron j to i.
Here, the weight and bias parameters are set as $w_{11}=-22.0, w_{12}=5.9, w_{21}=-6.6, w_{22}=0.0, \theta_1=-3.4$ and $\theta_2=3.8$ \cite{pasemann2002complex}, such that, if uncontrolled ($c_i^{(p)}(t)\equiv 0$) it shows chaotic dynamics. In order to obtain a given period p, the control input is given by:
\begin{equation}
	c_i^{(p)}(t) = \mu^{(p)}(t) \sum_{j=1}^{2}w_{ij}\Delta_j(t) \label{eqa2}
\end{equation}
It is calculated every $p$ time steps, while for the other steps it is set to 0. In Eq. \ref{eqa2}, $\Delta_j(t)$ indicates the activity difference between the current step and $p$-steps before:
\begin{equation}
	\Delta_j(t)=x_j(t)-x_j(t-p) \label{eqa3}
\end{equation}
and $\mu^{(p)}(t)$ is the control strength, which changes its value adaptively according to:
\begin{equation}
	\mu^{(p)}(t+1) = \mu^{(p)}(t) + \lambda \frac{\Delta^2_1(t)+\Delta^2_2(t)}{p}\label{eqa4}
\end{equation}
with an adaption rate $\lambda$, e.g. 0.05.

\begin{figure}
	\begin{center}
		\includegraphics[height=2.5in]{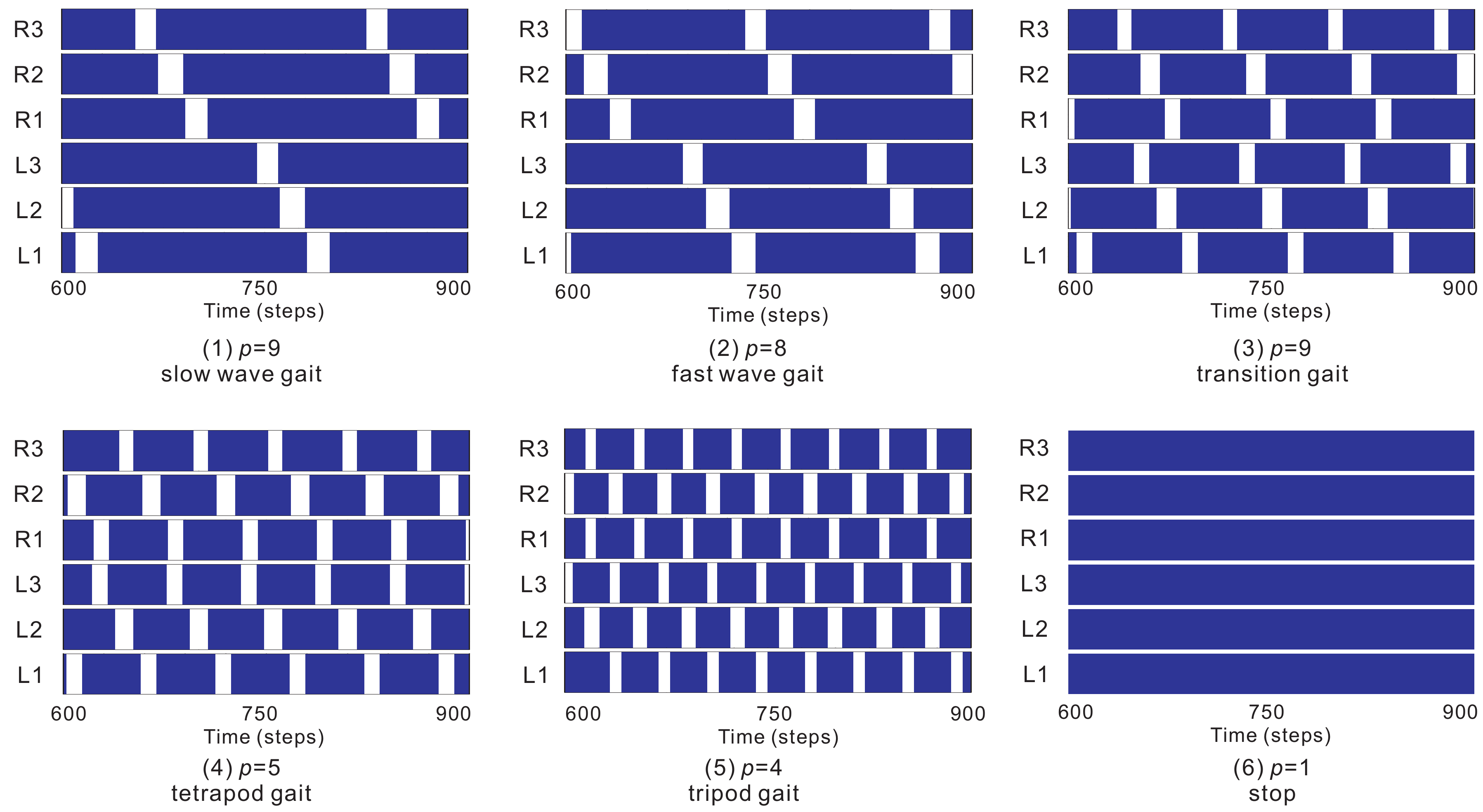}
	\caption{Different hexapod gaits for changing $p$ and the stop status ($p=1$). }
	\label{gaitfig}
	\end{center}
\end{figure}

Thus, using this single chaotic CPG, different oscillation periods can easily be obtained just by changing the $p$ value. After passing through some neural post-processing modules (see \cite{steingrube2010self} for details), different gait patterns are produced for a hexapod robot; with an increase of {\it p}, the robot walks slower. Periods 9, 8, 6, 5, 4 indicate slow wave gait, fast wave gait, transition gait, tetrapod gait and tripod gait, respectively (see Fig. \ref{gaitfig}). A blue area means that this leg is in a support phase, i.e., it touches the ground, while a white area indicates the swing phase. Note that one time step is $\approx$ 0.037 s. If $p=1$, there is no swing phase, i.e., the robot stops with all legs touching the ground. Periods 7 and 3 are unstable patterns while period 2 cannot generate a proper walking gait due to fast oscillation switching between two fixed points. Therefore, these periods are not used for locomotion generation.

\begin{figure}[!t]
\begin{center}
	\subfigure[Single CPG controller]{
      \label{6CPGs:a}
	\includegraphics[width=0.28\linewidth]{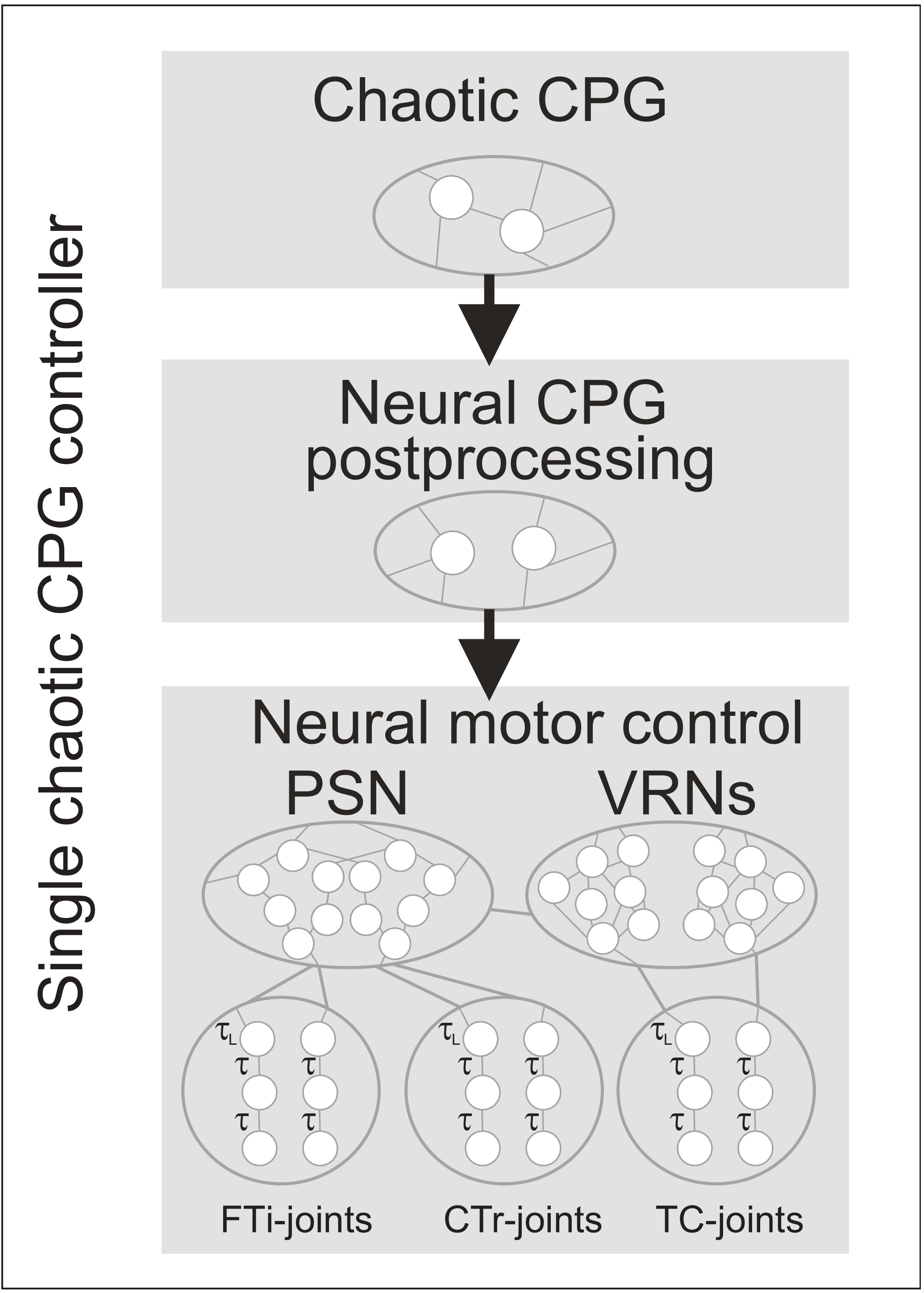}
	}
	\subfigure[Multiple CPGs controller]{
      \label{6CPGs:b}
	\includegraphics[width=0.6\linewidth]{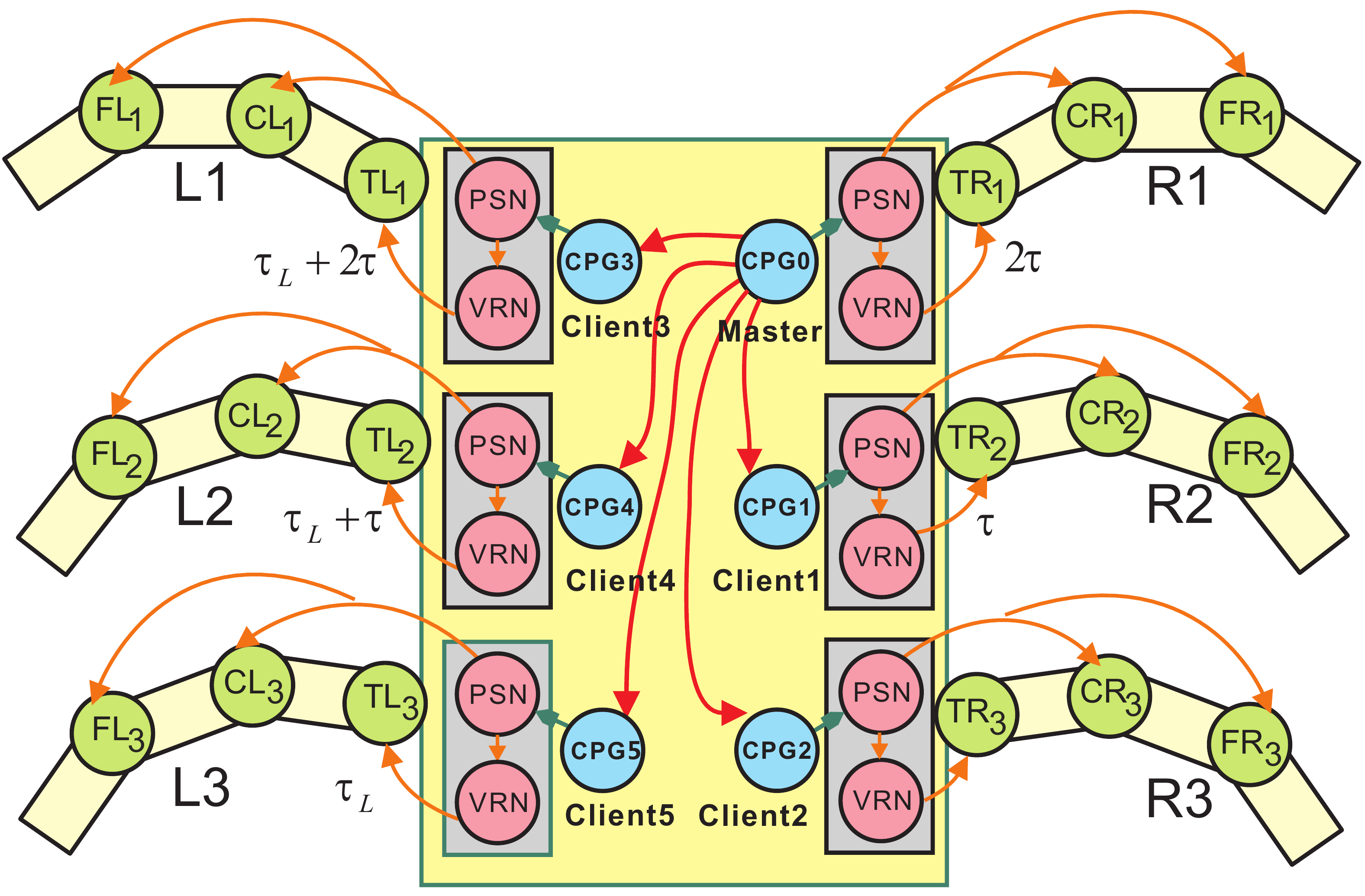}
	}
	\caption{Single chaotic CPG (a) and multiple chaotic CPGs (b) for a multi-legged robot. }
	\label{6CPGs}
\end{center}
\end{figure}

Another useful function of this mechanism is the chaotic output. If we set the control signal $c_i^{(p)}(t)\equiv 0$, the neural CPG circuit shows chaotic dynamics, which can be applied for self-untrapping, e.g., when a leg falls into a hole.
It is important to note that in principle chaotic dynamics is exploited not only for the self-untrapping but also mainly for obtaining a stable period from a large number of unstable periods embedded in chaotic dynamics. Thus, without chaotic dynamics, different stable periods cannot be obtained in this scheme.
Usually, the outputs of the CPG are passed to motor neurons, which activate the leg joints, through two hierarchical neural modules: neural CPG post processing and neural motor control (see Fig.~\ref{6CPGs:a} for an abstract diagram and \cite{steingrube2010self} for the complete neural circuit). Since the neural CPG post processing and motor control have already been presented in our previous studies \cite{manoonpong2008sensor,steingrube2010self}, we only discuss them briefly here.

The neural CPG postprocessing, which directly receives the CPG outputs, shapes the CPG signals to allow for smooth leg movements.
Subsequently, the post-processed CPG outputs are transmitted to the neural motor control module.
This module (see Fig.~\ref{6CPGs:a}) consists of one phase switching network (PSN) and two velocity regulating networks (VRNs). The PSN is a generic feed-forward network, which can reverse the phase of the processed CPG outputs with respect to a given input \cite{manoonpong2008sensor,steingrube2010self}. It is implemented to achieve a proper phase shift between the CTr- and FTi-joints and allow for sideward walking \cite{manoonpong2008sensor}. The VRNs are also simple feed-forward networks, with each VRN controlling the three ipsilateral TC-joints on one side. Since the VRNs act qualitatively like a multiplication function \cite{manoonpong07modular}, they have the capability to increase or decrease the amplitude of the TC-joint signals and even reverse them with respect to their control input. Controlling the TC-joint signal in this way results in various walking directions, such as forward/backward or turning left/right (see \cite{manoonpong2008sensor} for walking experiments). Finally, the outputs of the PSN and VRNs are sent to the motor neurons through delay lines. The ipsilateral lag is determined by a delay $\tau$ (i.e., 16 time steps or $\approx$ 0.6 s, see Fig.~\ref{6CPGs:a}) and the phase shift between both left and right sides are given by another delay $\tau_{L}$ (i.e., 48 time steps or $\approx$ 2 s, see Fig.~\ref{6CPGs:a}). This setup leads to biologically motivated leg coordination, since the legs on each side perform phase shifted waves of the same frequency \cite{wilson1966insect}. The frequency of the waves is defined by the period $p$ of the CPG, resulting in different gaits. In addition, sensory feedback can be integrated into the controller by targeting the VRNs; thereby leading to sensor-driven orienting behavior \cite{manoonpong2008sensor,manoonpong2007reactive}.


\subsection{Multiple chaotic CPGs}

The single chaotic CPG is able to handle most situations where all joints of the robot are functional. However, if one or more of the joints are disabled, the robot cannot use the same gait to stay on its trajectory. In other words, the robot cannot compensate for leg malfunction with a single CPG. In contrast, real insects can control their locomotion to continue with their trajectory even though some legs are malfunctioning or damaged \cite{graham1977effect}. To do this, their legs show different frequencies in order to obtain effective walking patterns.

Inspired by this, we extend the original single CPG to multiple CPGs where the other modules (i.e., CPG post processing, PSN and VRN networks) are also replicated at each leg.
The outputs of the PSN and VRNs of each leg are still sent to the corresponding motor neurons through the fixed delay lines, as described above. The control structure developed for hexapod locomotion is shown in Fig. \ref{6CPGs:b}. In this figure, the blue circles represent the CPGs. The right front neuron is the called master CPG, while other are called client CPGs. The neural motor control modules are depicted in gray rectangles. Each module consists of one PSN and one VRN which are depicted by pink circles. Green circles are motor neurons. The small dark green arrows indicate the signals from the CPG post processing modules (not shown here but see \cite{steingrube2010self}). Red lines indicate the synchronization mechanism. Orange lines indicate delay lines transmitting the outputs of the PSN and VRN to the motor neurons. After delaying the signals by corresponding time steps, the outputs of the motor neurons are sent to the leg joints.

The client CPGs can synchronize to the master in order to keep pace with the oscillation frequency. When synchronized, the controller generates the same outputs as if there was only one chaos control CPG. If some legs are disabled, the six CPGs automatically lose synchronization and can oscillate at different frequencies. A similar approach can be also applied to other walking schemes, e.g., quadruped locomotion (see below).

\subsection{Synchronization and desynchronization}

\begin{figure}
	\begin{center}
		\includegraphics[height=1.9in]{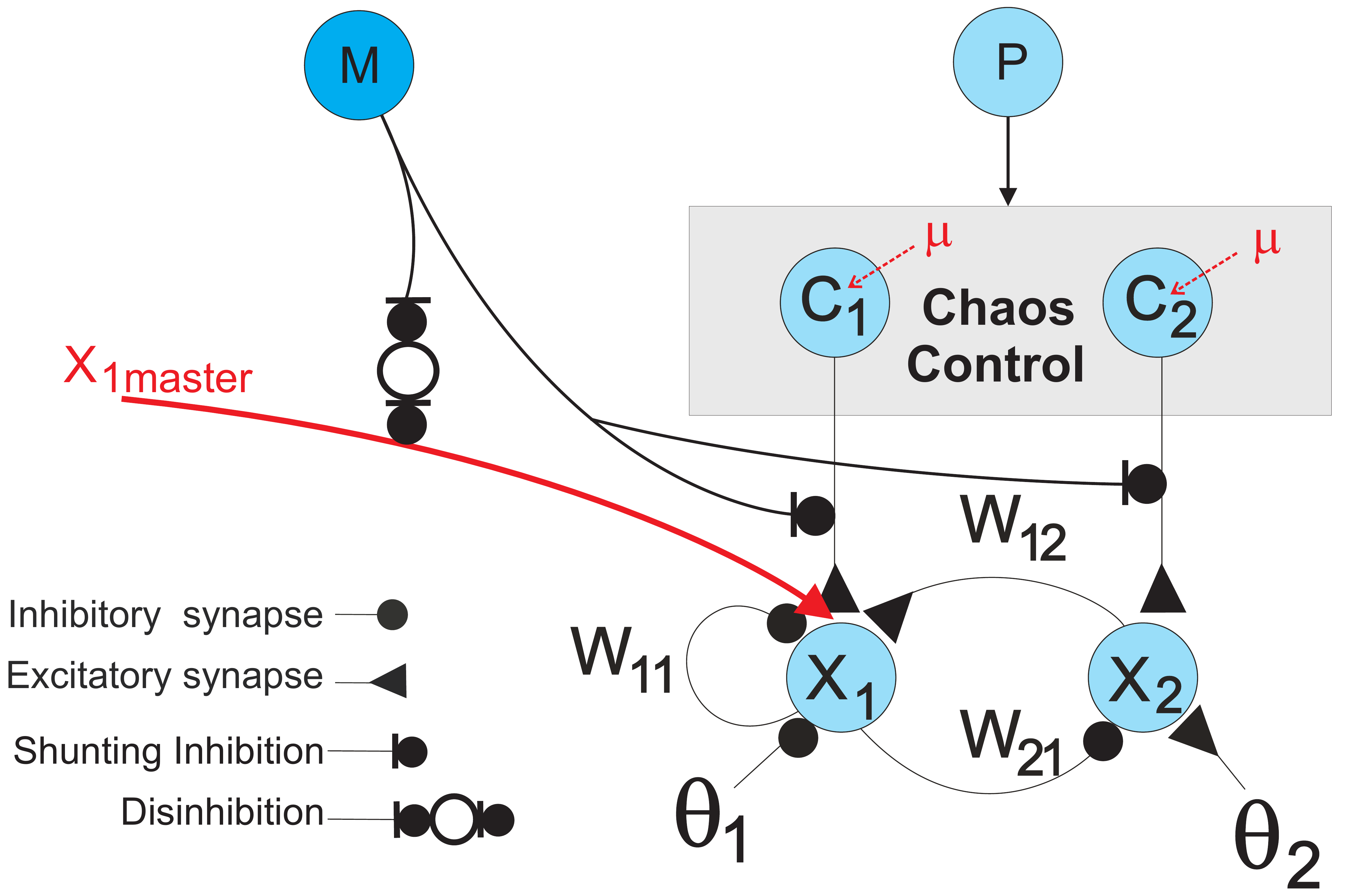}
	\caption{The inner structure of the client CPG.}
	\label{fig5}
	\end{center}
\end{figure}

The inner structure of the master CPG is like that of our original chaotic CPG (Fig.~\ref{fig3}), but the inner structure of the client CPGs is different. We add a synchronization mechanism to each client CPG as shown in Fig.~\ref{fig5}. It is similar to the master CPG except for the synchronization mechanism where the M-neuron is introduced to enable or disable the signal from the master CPG. When a client CPG needs to synchronize to the master CPG, the M-neuron becomes active (i.e., value of 1) shunting the synaptic weight from the inputs ($c_1, c_2$). Thus, the outputs of the network are uncontrolled and, the output from the master CPG ($X_{1master}$, see Fig.~\ref{fig5}) that was inhibited before is passed to the client due to disinhibition. This results in the output of the client-CPG oscillating at the same frequency as the master CPG. When the client CPG needs to oscillate at its own frequency, the M-neuron becomes inactive (i.e., 0) switching off the inhibition by cutting down the connection from the master CPG. The control inputs ($c_1$ and $c_2$) can again pass through to the network and, as a consequence, all the CPGs oscillate at different frequencies.

\begin{figure}[!t]
\begin{center}
	\subfigure[Synchronous]{
		\label{fig6:mini:subfig:a}
			\includegraphics[height=1.5in]{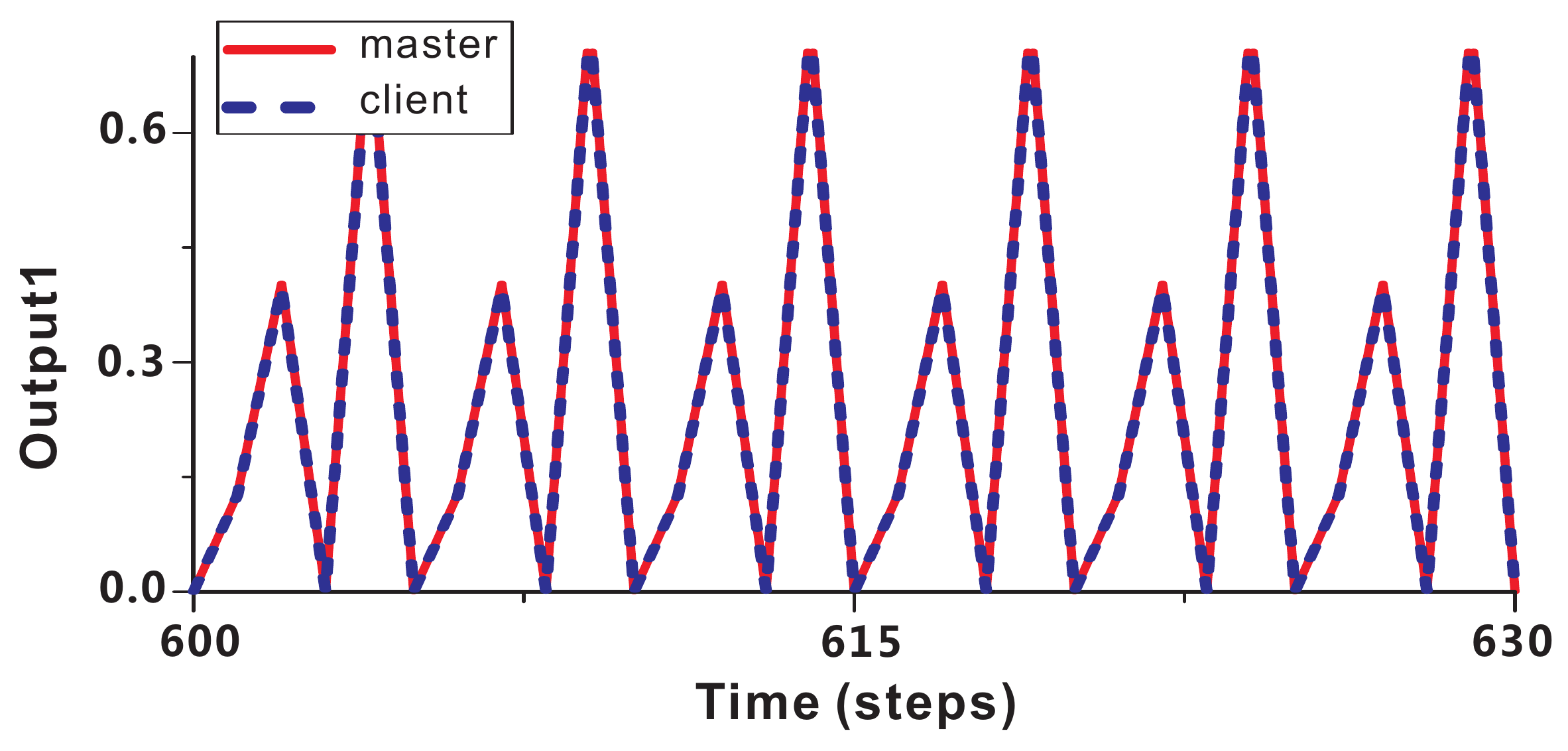}
	}
	\subfigure[Asynchronous]{
		\label{fig6:mini:subfig:b}
			\includegraphics[height=1.5in]{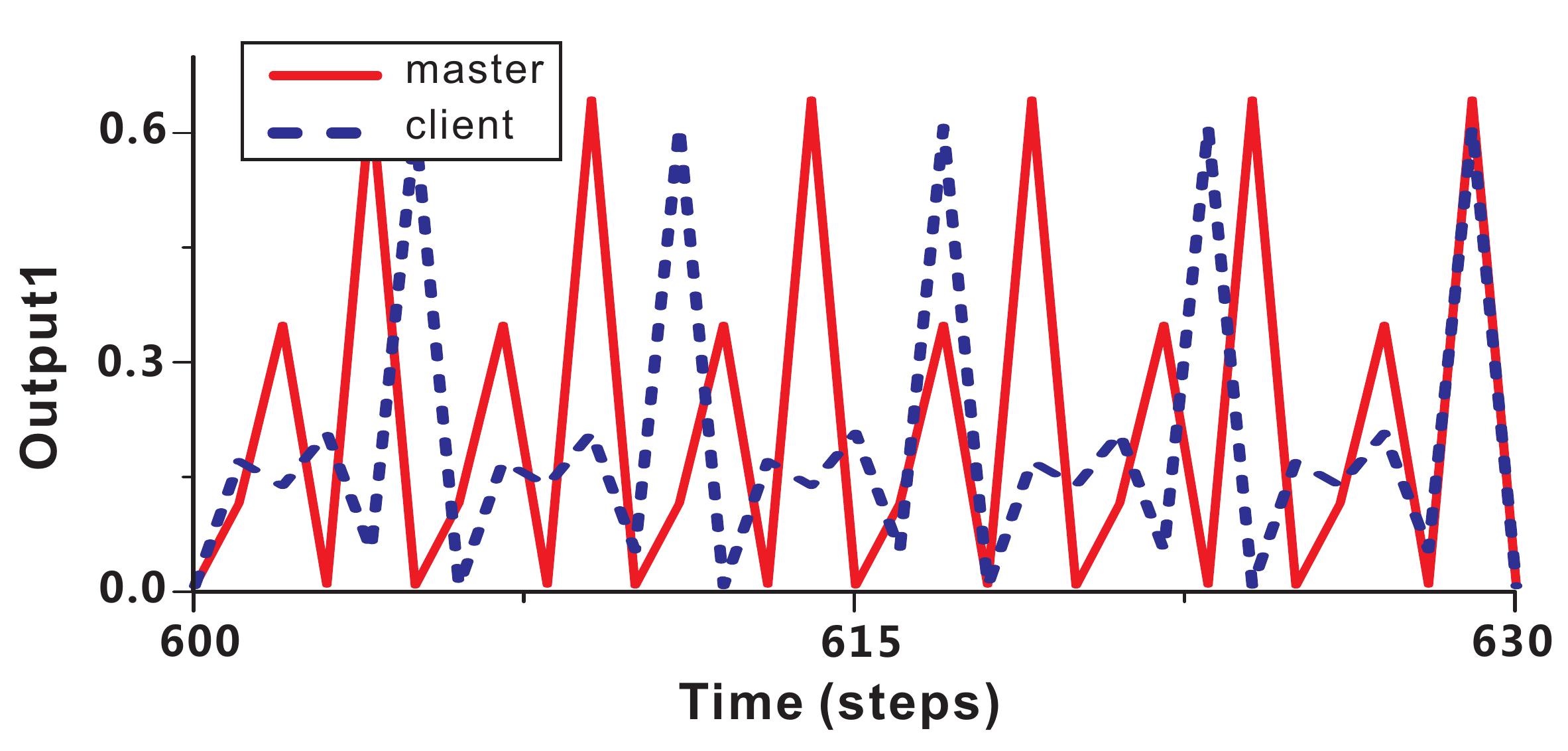}
	}
	\caption{The outputs of the CPG network for (a) synchrony and (b) asynchrony. }
	\label{fig6}
\end{center}
\end{figure}

The following equations describe the details of the client CPG.
With all legs functional, $\alpha$ is set to 1 such that the legs move with the same default frequency. If some legs malfunction, they are automatically set to 0 such that each leg can oscillate at its own frequency and learning will find a proper combination of oscillation frequencies of the different legs for malfunction compensation (see the next section).
The outputs of the two neurons satisfy:
\begin{equation}
\begin{array}{l}
	x_1(t+1) = \sigma(a_1(t)) + \alpha(x_{1master} -  \sigma(a_1(t)) ) \\
	x_2(t+1) = \sigma(a_2(t))
\end{array}
\label{eqa5}
\end{equation}
where the activity satisfies:
\begin{equation}
a_i(t) = \theta_i + \sum_{j=1}^{2}w_{ij}x_j(t) + c_i^{(p)}(t)
\label{eqa6}
\end{equation}
and $\alpha$ is the synchronization parameter. It is set to 1 if the M-neuron is active and 0 if it is inactive. The outputs of the first neuron for synchrony and asynchrony are shown in Fig.~\ref{fig6}. In this figure, the master CPG has period 5 and the client CPG has period 6. Note that one time step is $\approx$ 0.037 s.

\section{Learning for leg period adaptation}
\label{learn}

\begin{figure}
	\begin{center}
		\includegraphics[height=2.1in]{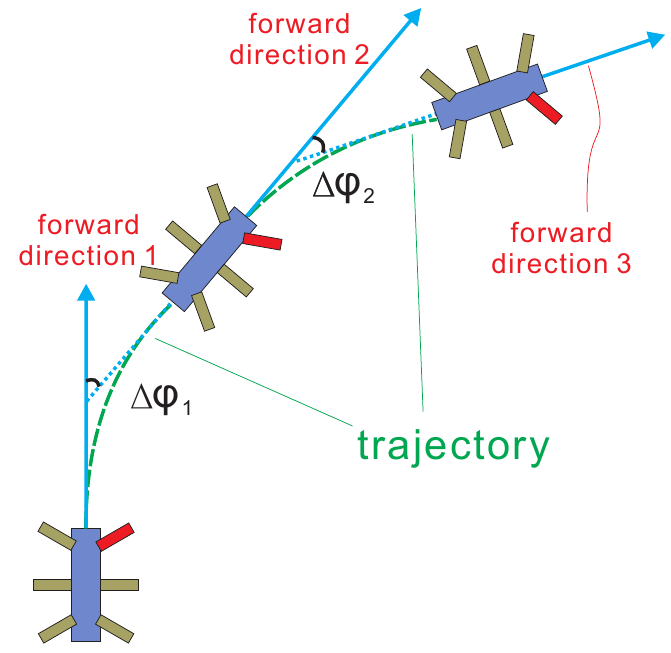}
	\caption{
The method of deviation measure in the learning algorithm. 
}
	\label{DeviaionMeasure}
	\end{center}
\end{figure}

The desynchronization of the multiple chaotic CPGs enables each leg to oscillate at its own frequency. To adapt the leg period automatically for malfunction compensation, here we apply simulated annealing \cite{dowsland1993simulated,bertsimas1993simulated} as our search algorithm to obtain feasible solutions. The algorithm is suitable to our task since it can be used for global optimization problems and discrete search space problems. The complete learning process is described as follows:

1) The robot starts to walk in a forward direction. While walking, its yaw angle is recorded to determine the heading direction of the robot. 2)  After a certain time window (400 time steps), the current yaw angle is measured and subtracted from the last or the initial one.
The angle difference $\Delta \varphi$ is used for estimating a deviation, which is illustrated in Fig.~\ref{DeviaionMeasure}. In red we depict the disabled leg. The green dashed line indicates the walking trajectory and the blue arrows point to the forward heading direction. $\Delta \varphi_1$ and $\Delta \varphi_2$ are the deviation angles, which are the yaw angle differences at certain steps. If there is no joint disabled, $\Delta \varphi$ should be approximately equal to zero, otherwise it indicates the degree of deviation. When a deviation occurs, i.e., when some joints are disabled, all  CPGs automatically lose synchronization and oscillate independently.
3) The oscillation period of each leg is stochastically changed, and the deviation is re-evaluated. 4) This process is repeated until the deviation is below a specific threshold and therefore leg malfunction compensation is achieved.

%

\begin{table}[!t]
\caption{Learning algorithm.}
\label{tabflowchart}
\centering
\begin{tabular}{l}
\toprule[2pt]
{\bf Initialize} $C(1) = [4,4,4,4,4,4]; \Delta\varphi = 0.0; E_1=0.0$ \\
{\bf Repeat:}\\
At trials $n$ \\
(1) randomly pick a leg $l$, $l \in [R1, R2, R3, L1, L2, L3]$ \\
(2) change the period of leg $l$ to a random value, $P(l) \in [4, 5, 6, 8, 9]$\\
(3) compare this combination of leg period, $C'(n)$, to the walking records\\
(4) repeat (1) to (3) until $C'(n)$ is a new combination of leg periods \\
(5) run the robot \\
(6) calculate the evaluation function and its variation \\
\multicolumn{1}{c}{$E_n = \Delta \varphi$} \\
\multicolumn{1}{c}{$\Delta E = E_n - E_{n-1}$} \\
(7) decide the combination of leg periods \\
 \qquad \qquad \qquad \qquad \qquad {\bf if} $\Delta E < 0$ {\bf then} \\
 \qquad \qquad \qquad \qquad \qquad \qquad $C(n) = C'(n)$ \\
 \qquad \qquad \qquad \qquad \qquad{\bf else} \\
 \qquad \qquad \qquad \qquad \qquad \qquad{\bf if} $x \leq e^{-\beta \Delta E}$ {\bf then} \\
 \qquad \qquad \qquad \qquad \qquad \qquad \qquad$C(n) = C'(n)$ \\
 \qquad \qquad \qquad \qquad \qquad \qquad{\bf else} \\
 \qquad \qquad \qquad \qquad \qquad \qquad \qquad$C(n) = C(n-1)$ \\
 \qquad \qquad \qquad \qquad \qquad \qquad{\bf end if} \\
 \qquad \qquad \qquad \qquad \qquad{\bf end if} \\
{\bf Until:} The evaluation function $E_n$ is less than a required value $E_{req}$ \\
\bottomrule[2pt] 
\end{tabular}
\end{table}

Table.~\ref{tabflowchart} depicts the learning process. 
In this table, $E_{1,2,...,n}$ denote the evaluation function in different trial.
The function is calculated from the deviation angle, $\Delta \varphi$.
$E_{req}$ indicates the required minimum evaluation function. 
$\Delta E = E_n - E_{n-1}$ is the difference of the evaluation function between two trials.
$\beta$ is the adaption rate.
$X$ is a random value between 0 and 1.
$C'(n)$ and $C(n)$ indicate the randomly selected combination of leg periods and the chosen combination of leg periods in the $n^{th}$ trial, respectively.
If a deviation which is larger than a threshold occurs, the learning loop starts.
A random leg is selected and one of the five oscillation periods (period 4, 5, 6, 8, 9) are randomly assigned to this leg.
Then the combination of periods is compared to the previous states.
If this particular combination of periods has already been performed, the trial is aborted, and another random leg is selected and randomly assigned to another period. Once a combination has been selected, the robot moves forward and the deviation angle $\Delta \varphi$ is measured. We set $E_n = \Delta \varphi$; if $E_n$ is less than our required evaluation $E_{req}$, which means the deviation has already been compensated, the learning process is stopped and the loop is exited.
If $E_n$ is less than the evaluation of the last trial, i.e., $\Delta E = E_n - E_{n-1} < 0$, which means the deviation angle in this trial is less than last time, this combination of periods is stored and the next trial starts.
If $\Delta E> 0$, this combination of periods is stored with probability $X<e^{-\beta \Delta E}$, which is in the range from 0 to 1 and decreases as $\Delta E$ increases. For example, if $\Delta E$ is very large, which means the deviation is much larger than last time, the acceptance probability $ e^{-\beta \Delta E}$ is very close to 0. Thus, it is much more probable to return to the combination of periods of the last trial. Conversely, if the deviation is very small, it is more probable to keep the new combination of periods, and a new loop will start based on this new combination of periods.
The learning loop will be exited once the robot can walk straight even with disabled joints.
With this learning process, the robot can automatically learn to find a feasible combination of its leg periods. In this implementation, the search algorithm is conducted continuously until a suitable solution is found.

In the learning process, $\beta$ is an important parameter which creates a tradeoff between acceptance probability and convergence speed (small $\beta \rightarrow $ higher acceptance probability \& slower learning and vice versa). We additionally observe that for large $\beta$ the learning algorithm will often find inappropriate solutions. In this work, $\beta$  is selected empirically (usually to 0.5) to balance this trade off. This, together with the use of a cost function (the ``deviation angle") and a small, finite set of leg periods assures convergence of the simulated annealing method \cite{bertsimas1993simulated}.

\section{Implementation on different walking robots}
\label{sect4}
\subsection{Simulated six legs (6 CPGs)}

\begin{figure}
	\begin{center}
		\includegraphics[height=2.0in]{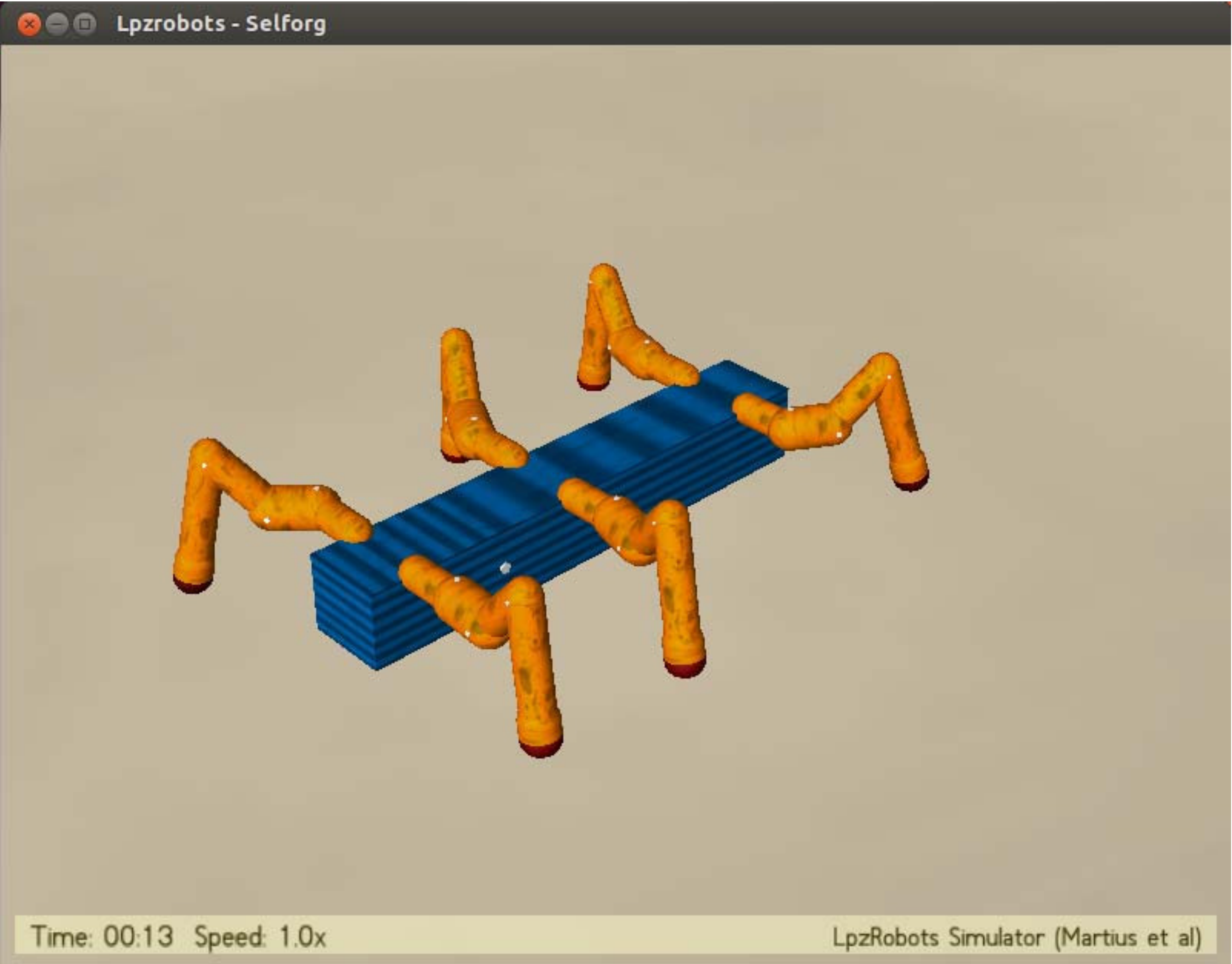}
	\caption{A hexapod robot in LPZROBOTS.}
	\label{hexa}
	\end{center}
\end{figure}

LPZROBOTS\footnote{It is based on the Open Dynamics Engine (ODE). For more details of the LPZROBOTS simulator, see \url{http://robot.informatik.uni-leipzig.de/software/}.} was employed as a simulation environment. For testing our multiple CPGs and the learning algorithm, a six CPGs controller was implemented on a simulated hexapod robot as shown in Fig.~\ref{hexa}. The controller is updated with a frequency of 27Hz. With all the legs initially working well, we switched on the M-neuron to synchronize the client-CPGs to the master CPG. At the same time, the master CPG was set to periods 4, 5, 6, 8, 9, respectively (see. Fig.~\ref{gaitfig}). All legs performed at the same frequency and achieved tripod gait, tetrapod gait, transition gait, fast wave gait and slow wave gait, respectively. In synchrony, the robot performed just as if there was only one CPG.

To simulate leg malfunction, we disabled the movement of one or more legs by setting the outputs of the three joints of the leg to constant values. As a consequence, the affected leg could not move normally but only sustain part of the body weight intermittently (depending on the other legs' status).
After disabling, the robot could not stay on a straight trajectory.
In the simulation, an orientation sensor was implemented on the robot to measure its yaw angle. If the orientation sensor detected a deviation, the six CPGs automatically lost their synchronization and oscillate independently. Afterward, the learning process began as described in Section~\ref{learn}. For every 400 time steps, we calculate the deviation angle by subtracting yaw at the start from the end of this time window. This deviation was evaluated against the current combination of leg periods and the robot tested different combinations until a suitable one is found. As a result, leg malfunction was compensated for, and the robot maintained its forward walking in a straight line. It is important to note that in our implementation the disabled leg was not taken into account for period change since changing its period does not affect the walking behavior. Thus, its period was kept fixed.

\begin{figure}
	\begin{center}
		\includegraphics[height=2in]{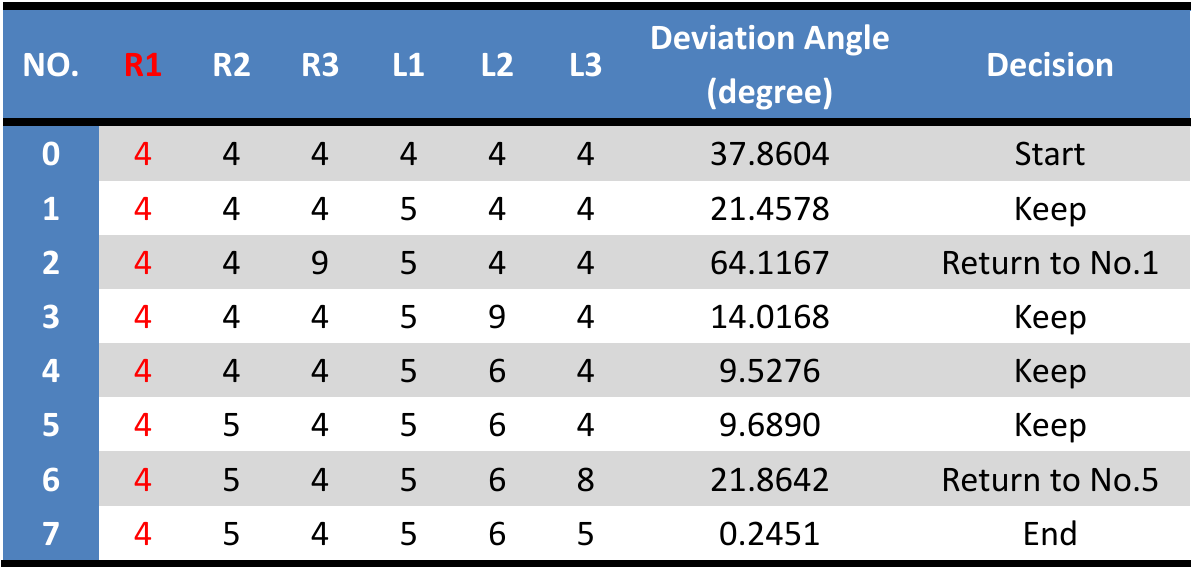}
	\caption{The learning process for one scenario (see text for details).}
	\label{pchange}
	\end{center}
\end{figure}

As an example, Fig.~\ref{pchange} illustrates the learning process for one scenario. In this figure, the first column indicates the $n^{th}$ trial. R1 = right front, R2 = right middle, R3 = right hind, L1 = left front, L2 = left middle, L3 = left hind. These six columns show the periods of the corresponding legs. The deviation in this trial is shown in degree. The last column shows the decision to keep this new period combination or return to a former one. In this scenario, the R1 leg was disabled (depicted in red) and its period was ignored in the learning process. Initially, the robot walked with tripod gait, i.e., every CPG oscillated with period 4. When the right front leg (R1) was disabled, the robot performed a right turning curve since it cannot provide enough propelling force to balance the body. After 400 time steps, the robot deviated to the right with an angle of deviation $\approx 37.86^{\circ}$. In step 1, the left front leg (L1) was randomly selected and randomly changed to period 5. After 400 time steps, the deviation angle was $\approx 21.46^{\circ}$. This combination of periods was kept since the deviation angle decreased. In step 2, the right hind leg (R3) was selected and changed to period 9 and the deviation was $64.12^{\circ}$. As $\beta = 0.5$, the acceptance probability of this combination of periods satisfies:
\begin{equation}
P= e^{-\beta \Delta E} = e^{- 0.5 \times (64.12-21.46)} = 5.45 \times 10^{-10},
\label{eqa7}
\end{equation}
which is very small. As a result, this trial was aborted and the robot returned to step 1. Hence, step 3 was derived from step 1 rather than step 2.  The deviations in step 3 and step 4 decreased so they were kept. In step 5, the deviation was a little larger than in step 4. However, the probability of keeping this combination of periods was close to 1, and it was kept. This shows the advantage of the simulated annealing method: it accepts worse situations which provide an opportunity to approach a better solution. Step 6 was aborted. In step 7, we obtained a final solution through changing the left hind leg (L3) from period 4 (in step 5) to period 5. Final deviation was only $\approx 0.25^{\circ}$.

\begin{figure}
\begin{center}
	\subfigure[Leg R1: disabled]{
		\label{jointangle:mini:subfig:a}
			\includegraphics[width=0.3\linewidth]{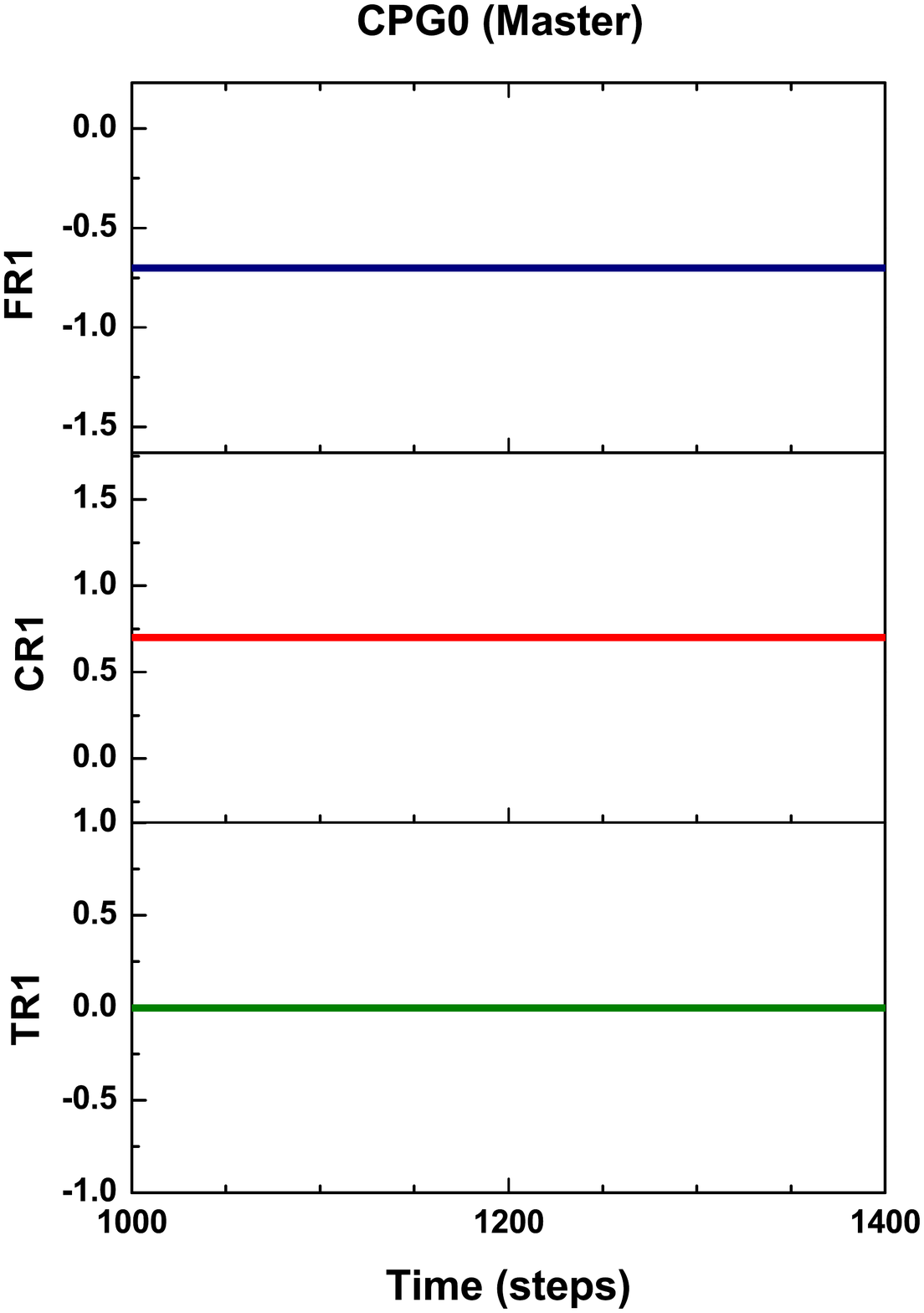}
	}
	\subfigure[Leg R2: p5]{
		\label{jointangle:mini:subfig:b}
			\includegraphics[width=0.3\linewidth]{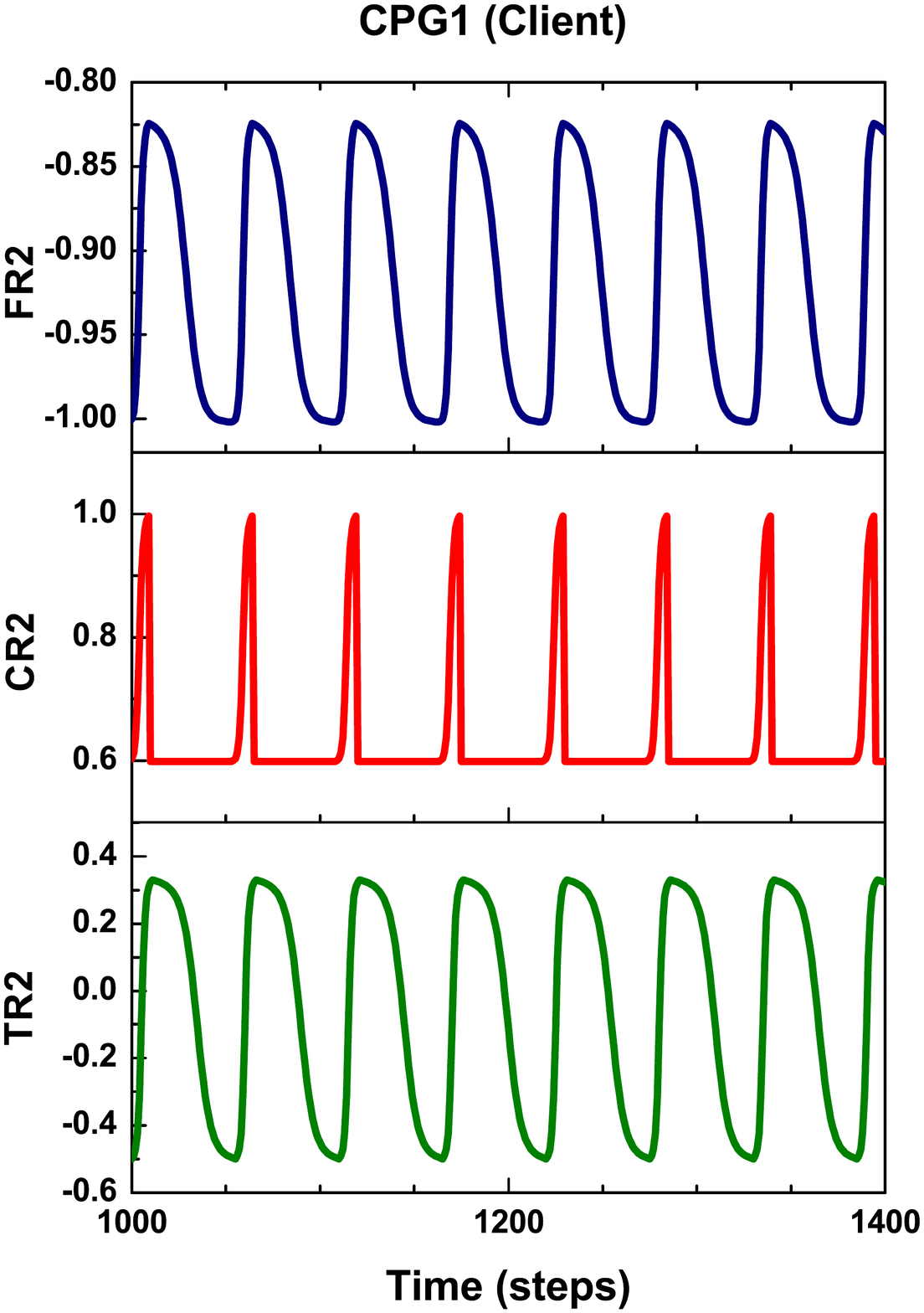}
	}
	\subfigure[Leg R3: p4]{
		\label{jointangle:mini:subfig:c}
			\includegraphics[width=0.3\linewidth]{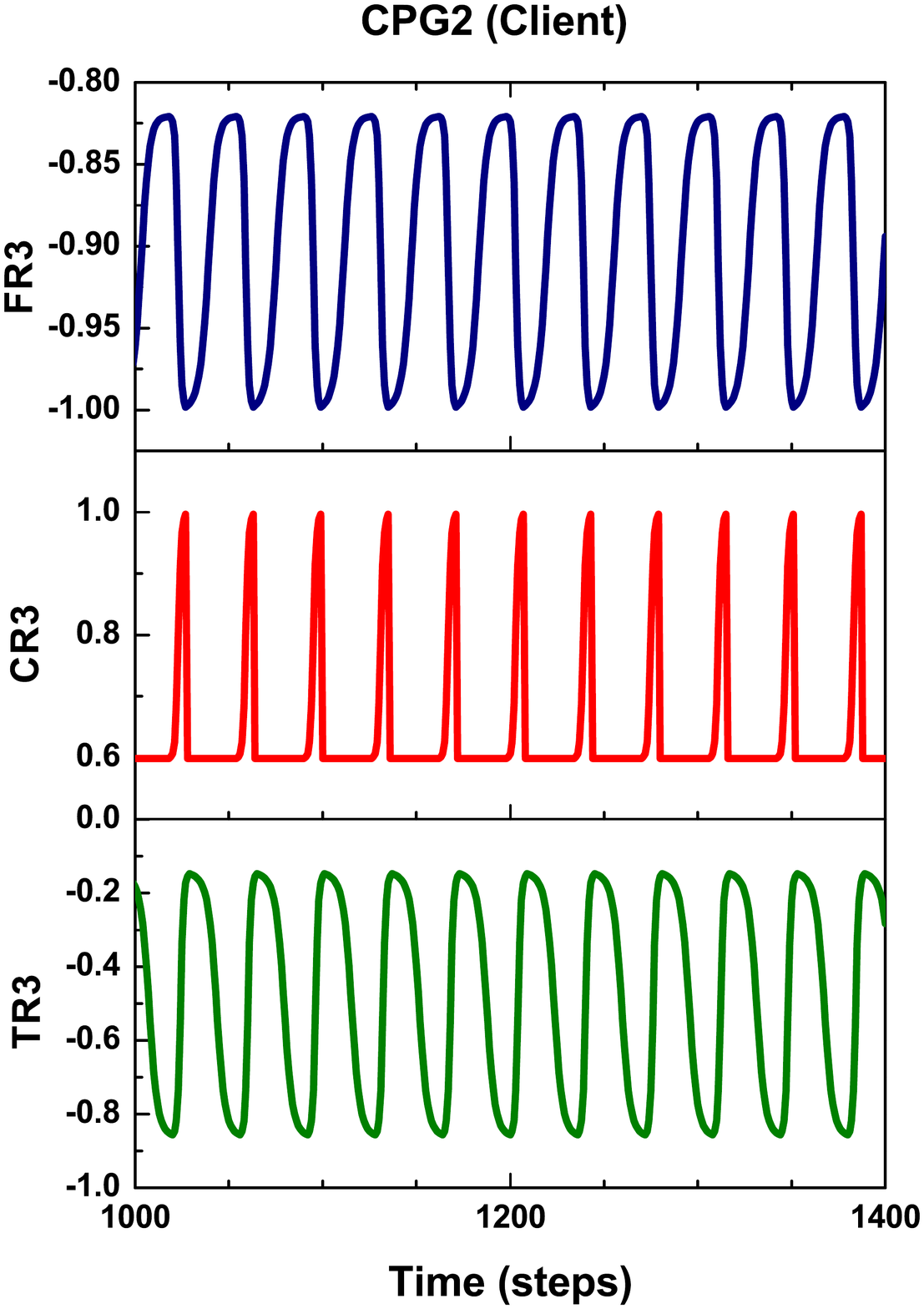}
	}\\
	\subfigure[Leg L1: p5]{
		\label{jointangle:mini:subfig:d}
			\includegraphics[width=0.3\linewidth]{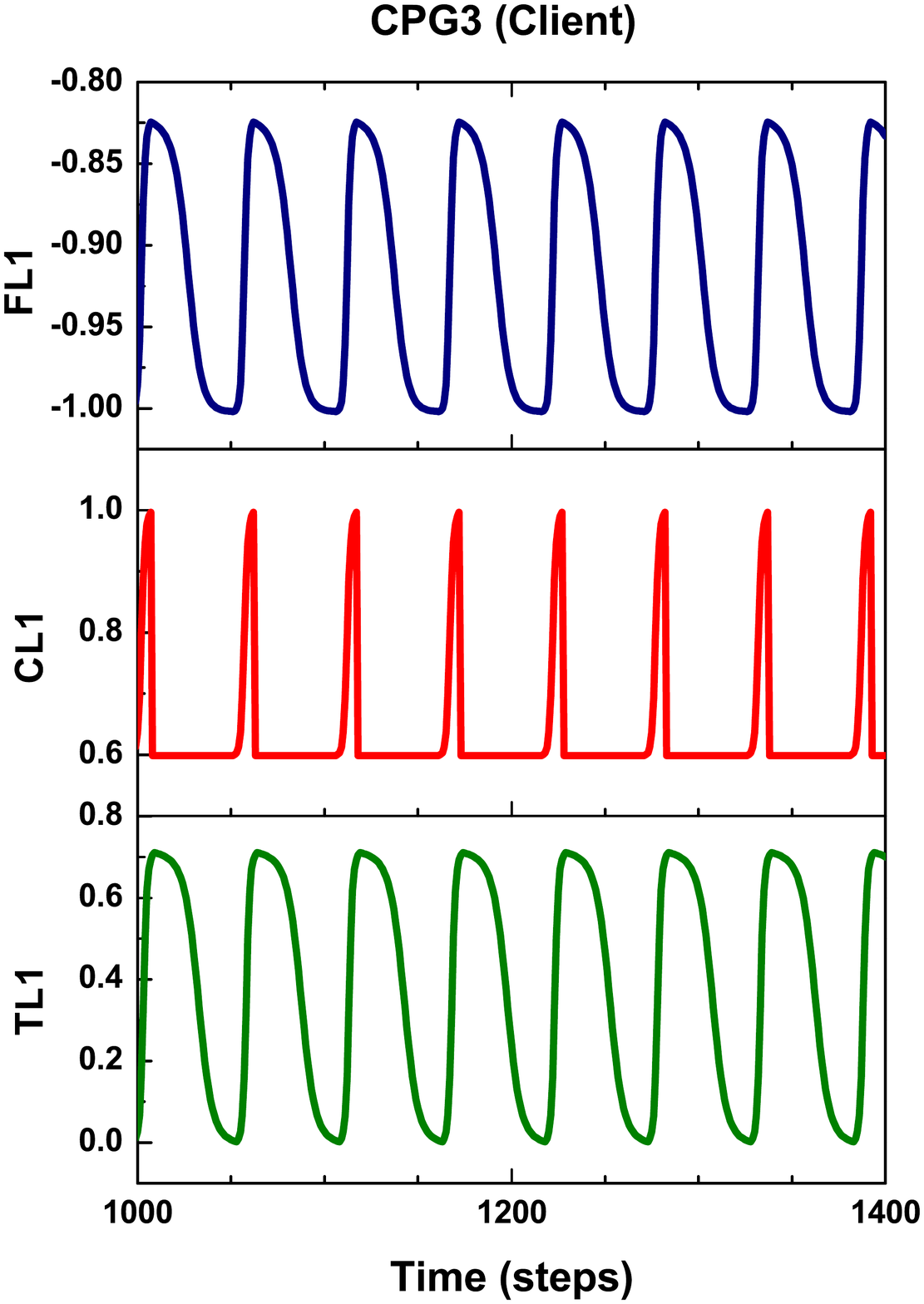}
	}
	\subfigure[Leg L2: p6]{
		\label{jointangle:mini:subfig:e}
			\includegraphics[width=0.3\linewidth]{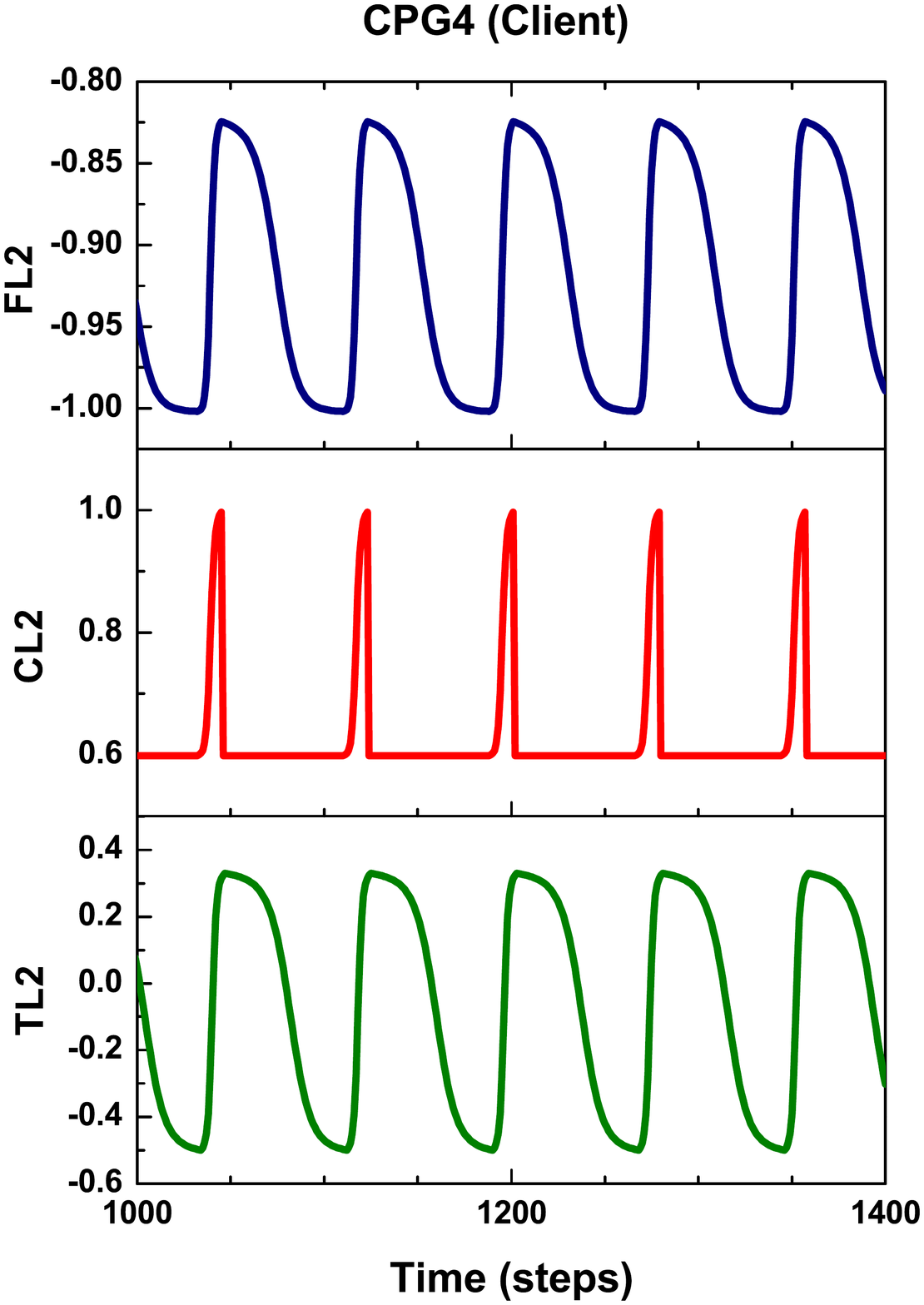}
	}
	\subfigure[Leg L3: p5]{
		\label{jointangle:mini:subfig:f}
			\includegraphics[width=0.3\linewidth]{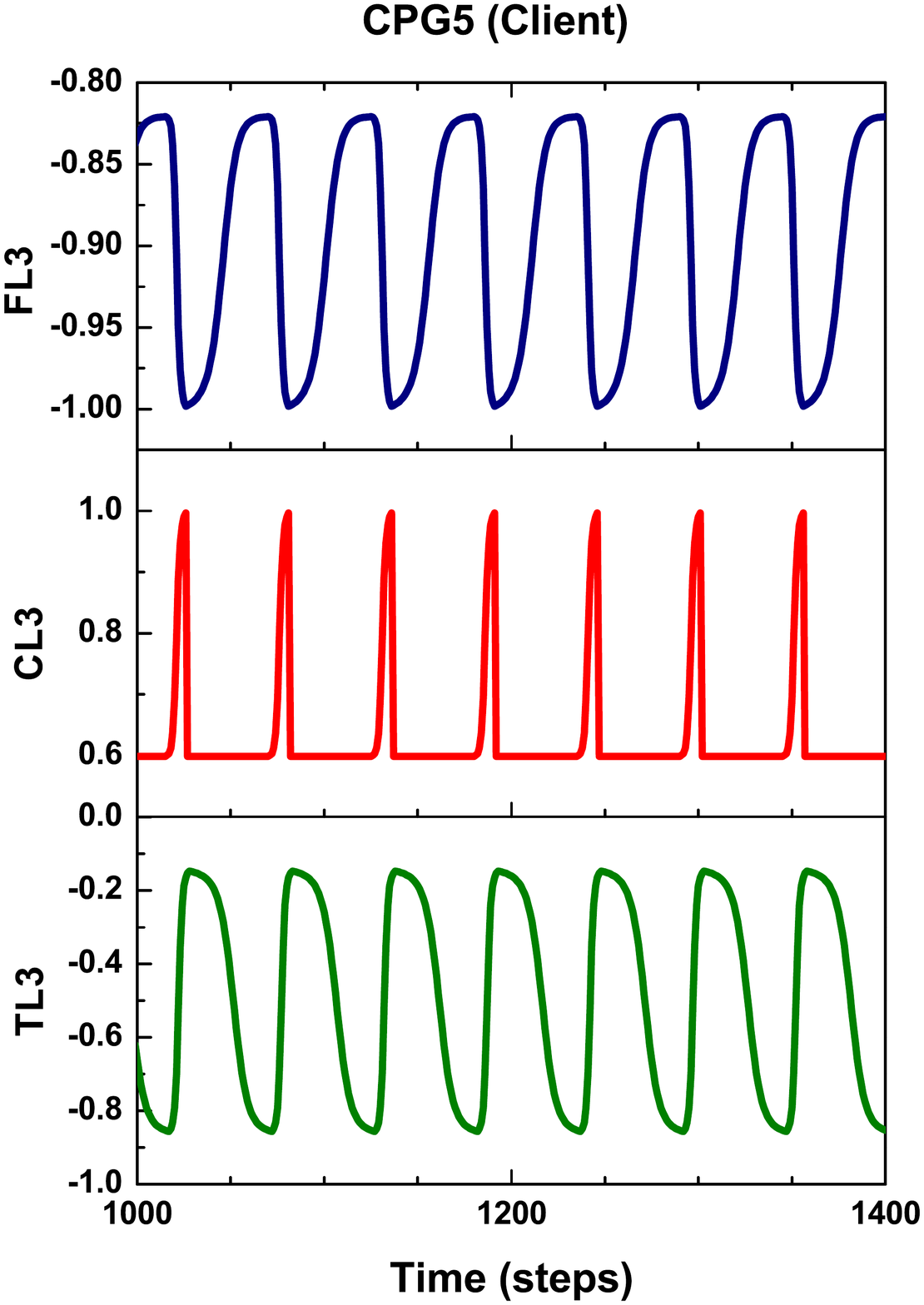}
	}
	\caption{Outputs of the 18 motor neurons. The FTi-, CTr- and TC- joints are depicted
in blue, red and green, respectively. Leg R1 (a) was disabled (i.e., the outputs of the three joints were kept at a constant value) while L2 (e) oscillated with period 6. 
The oscillation of R2 (b), L1 (d) and L3 (f) were decreased to period 5 and R3 (c) stayed on period 4. }
	\label{jointangle}
\end{center}
\end{figure}

According to this, a possible solution to cope with the problem of the R1 leg malfunction was a combination of the following periods: R2 = 5, R3 = 4, L1 = 5, L2 = 6 and L3 = 5. The joint angles of the 18 actuators performed as shown in Fig.~\ref{jointangle}. 
In this figure, the outputs of the FTi- joints of hind legs (e.g., Leg L3 and R3) were reversed in order to push the leg outwards to stabilize the body. Note that one time step is $\approx$ 0.037 s. 

\begin{figure}[!t]
	\begin{center}
		\includegraphics[height=4in]{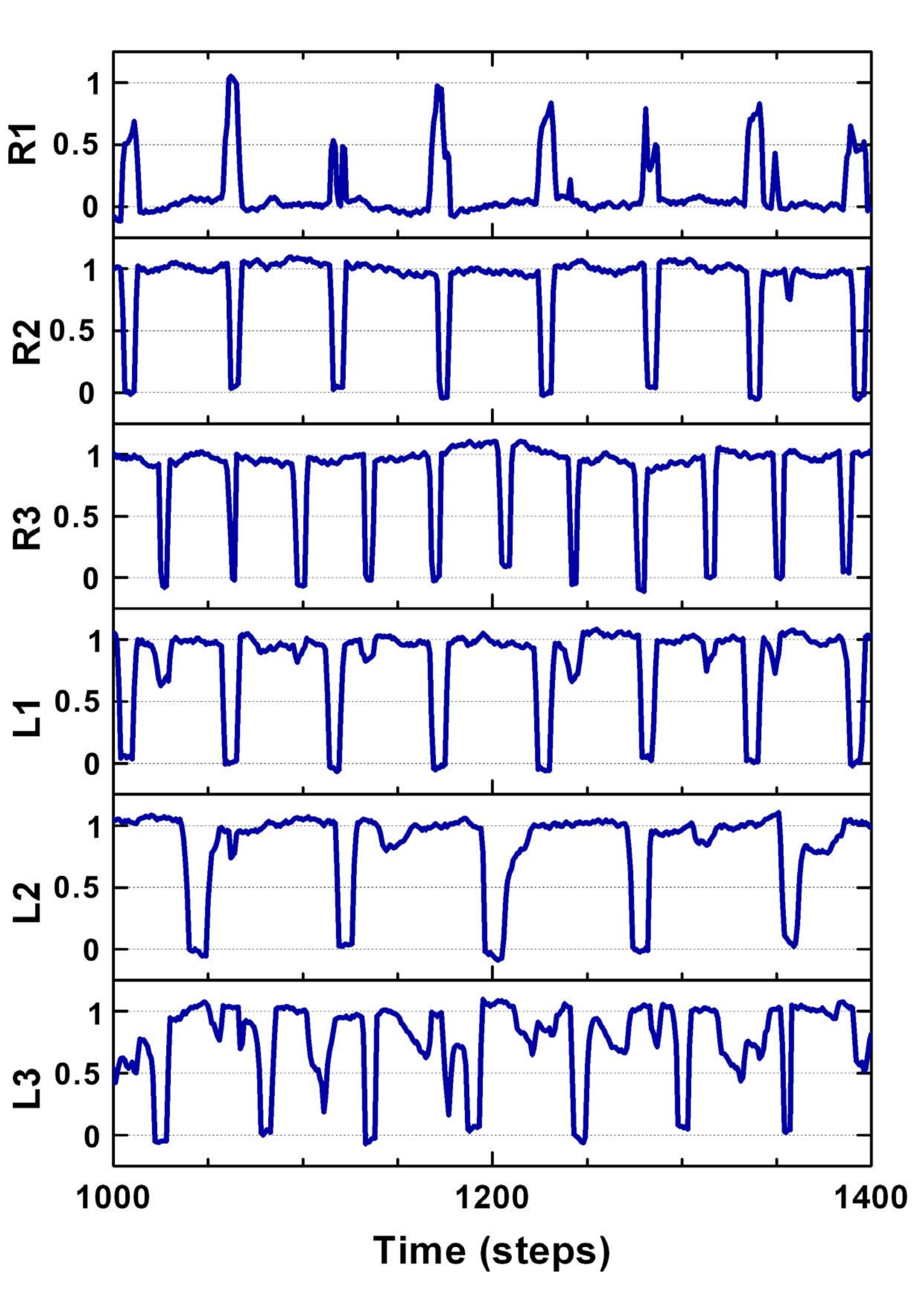}
	\caption{The foot contact signals of the six legs. 
Here, the leg R1 was disabled.
}
	\label{force}
	\end{center}
\end{figure}

The foot contact forces were recorded to see how the robot interacts with the environment (see Fig.~\ref{force}). In this figure, the signals have a range from 0 to 1. Positive 1 means the leg fully touches the ground (stance phase), while negative 1 means the leg is in the air (swing phase). The functional legs performed similar patterns relating to the input oscillation: for example, the R3 leg touched the ground with p4 and the L2 leg touched the ground with p6. However, the disabled R1 leg still touched the ground intermittently to support the body, especially when the neighboring legs attempted to swing (e.g., R2 and L1), even though its joints were fixed.

\begin{figure}
	\begin{center}
		\includegraphics[height=6.2in]{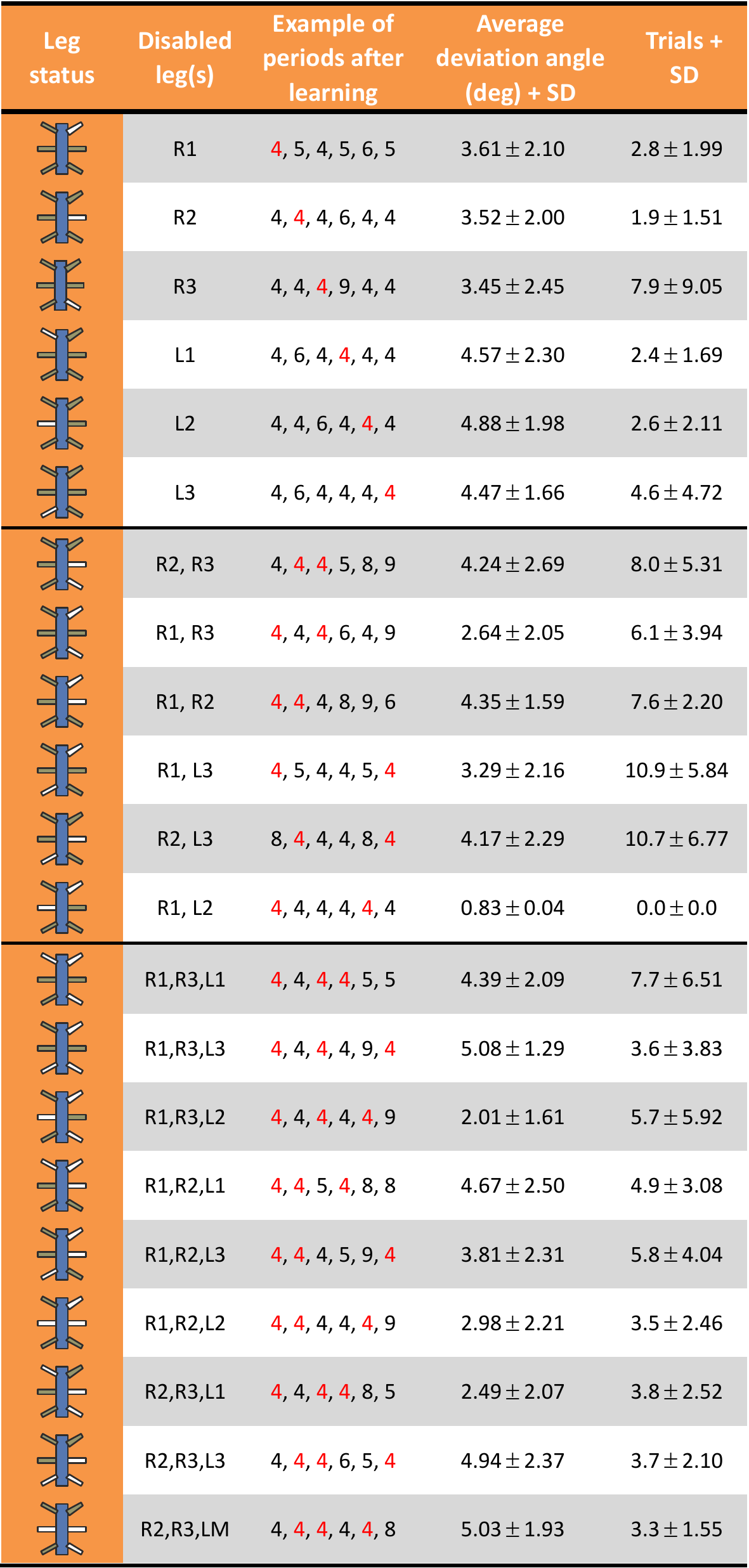}
	\caption{Combination of periods after learning for the hexapod. The first two columns illustrate the functional and disabled legs. The third column is the combination of periods after learning. In red we depict the disabled legs. The last two columns represent the deviation angle and how many trials are on average required with standard deviations (SD).
}
	\label{learningresult}
	\end{center}
\end{figure}

Other scenarios for learning to compensate leg malfunction were also tested and the results are depicted in Fig.~\ref{learningresult}. 
For each scenario, 10 trials had been averaged. In this diagram, all simulations started from period 4. In all cases, the learning state space for the hexapod consists of $5^6 = 15625$ different controllers.
In this figure, six scenarios for one leg's malfunction, six scenarios for two legs' malfunction and nine scenarios for three legs' malfunction were tested. 
These scenarios include most of the situations that might occur (see the first two columns), excluding situations that have a symmetrical counterpart in this Table. For example, the situation with the L2 and L3 legs disabled is not depicted here because we can deduce it from the scenario that has the R2 and R3 legs disabled. In Fig.~\ref{learningresult}, the best resulting period combination of each scenario is presented.

The actual time needed to conduct a real robot experiment can be approximately estimated. As one trial takes 400 time steps and the control frequency for a robot experiment is 27 Hz, it costs $400/27 \approx 14.8$ seconds to try one combination of periods. From the results shown in Fig.~\ref{learningresult}, the learning time for most of the scenarios can be normally controlled within 20 trials, i.e., it takes about 4 to 5 minutes to finish one learning experiment.

\subsection{Simulated four legs (4 CPGs)}

\begin{figure}
	\begin{center}
		\includegraphics[height=2in]{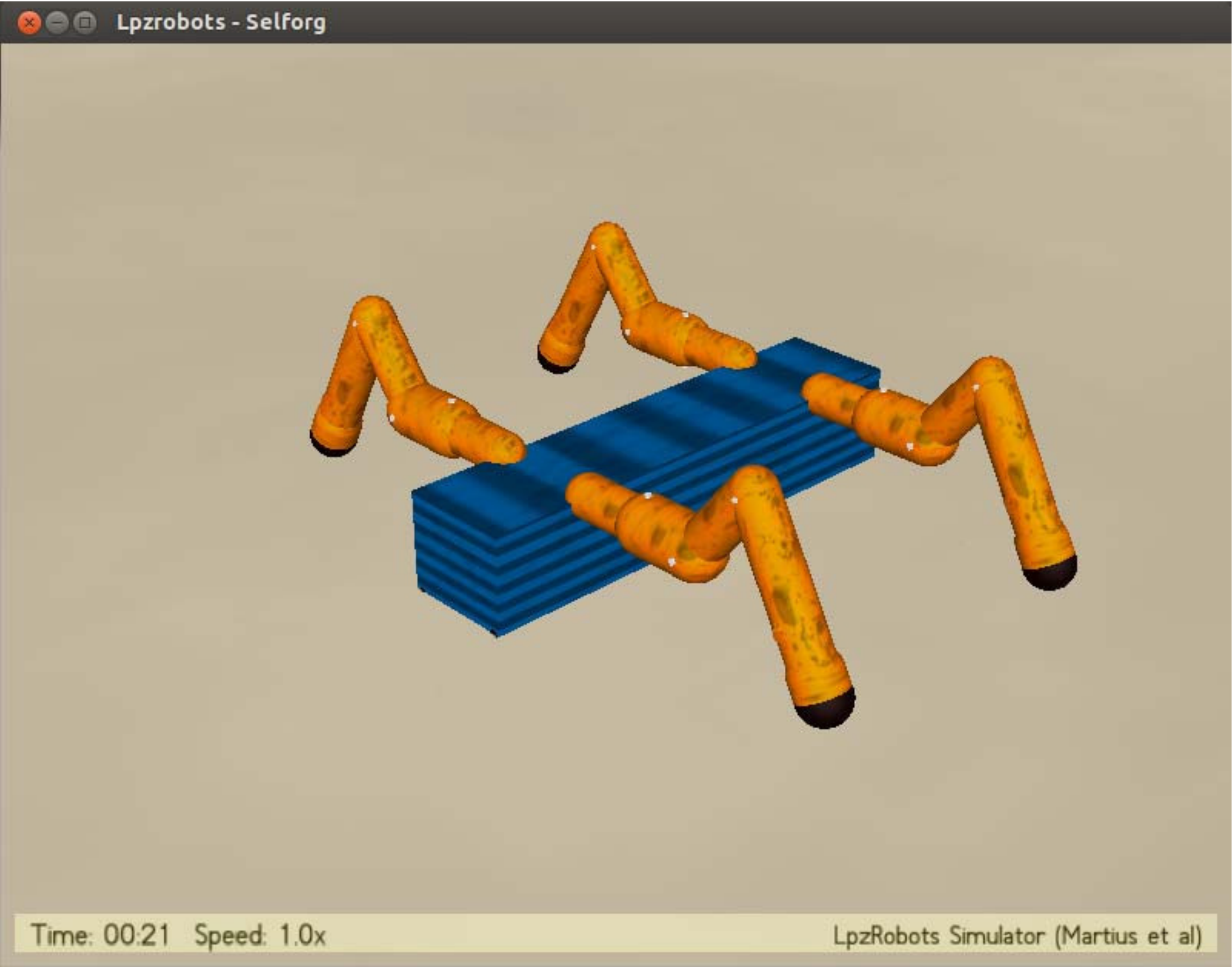}
	\caption{A simulated quadruped robot in LPZROBOTS.}
	\label{fourlegsimu}
	\end{center}
\end{figure}

\begin{figure}
	\begin{center}
		\includegraphics[height=2in]{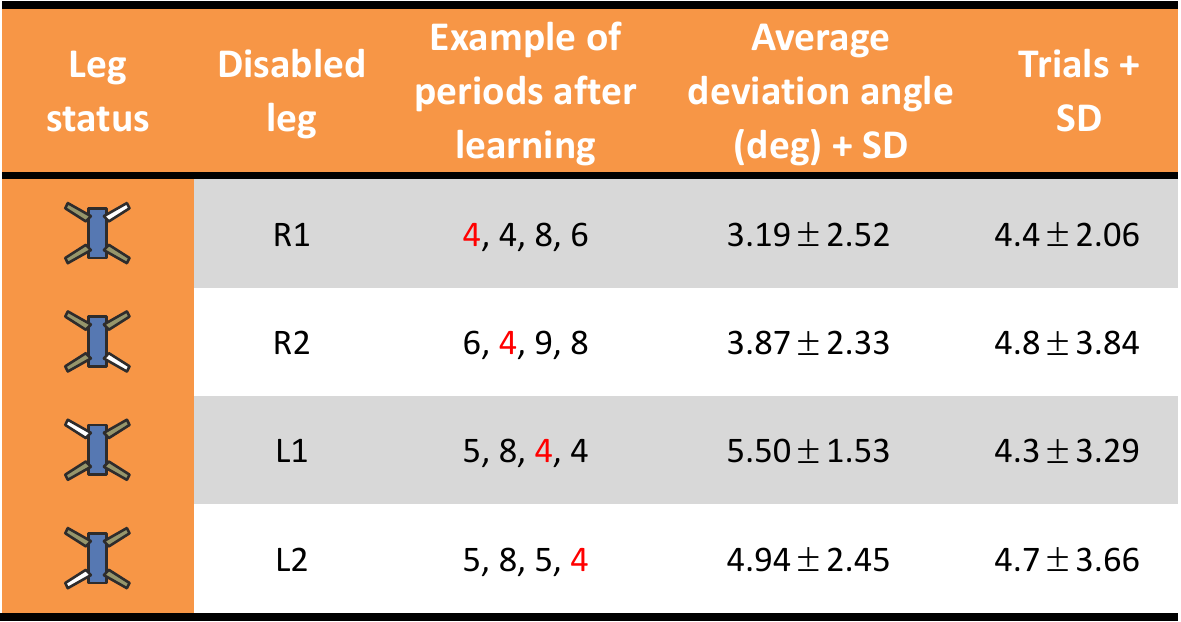}
	\caption{Learning results for quadruped locomotion. All columns are presented similar to Fig.~\ref{learningresult}.}
	\label{fourleglearn}
	\end{center}
\end{figure}

The multiple chaotic CPGs and the learning algorithm presented here are not limited to hexapod robots but can also be applied to different walking robots. To verify this, we simulated a quadruped robot in LPZROBOTS (see Fig.~\ref{fourlegsimu}).
For the controller of this robot, we used 4 CPGs with, as before, one CPG set as master CPG and the other three as clients. The controller is updated with a frequency of  27Hz. Synchronization and learning mechanisms were the same as for the 6 CPGs' controller. All trials started from period 4, i.e., a trot gait for the quadruped robot. In this simulation, we disabled the R1, R2, L1 and L2 legs individually to test the learning process for leg malfunction compensation (see Fig.~\ref{fourleglearn}). 
Similar to Fig.~\ref{learningresult}, the average deviation angles (deg) and trials are shown together with the standard deviations (SD). R1 = right front, R2 = right hind, L1 = left front, L2 = left hind. Every scenario was tested 10 times. In all cases, the learning state space for the quadruped consists of $5^4 = 625$ different controllers.
The experimental results show that the controller can find feasible combinations of periods in the different conditions of leg malfunction for the quadruped robot. This shows that our multiple CPGs and learning mechanism can be applied to not only a hexapod robot but also to a quadruped robot, indicating a certain potential for generalization. In other words, this generic approach can deal with not only small (i.e., quadruped) but also large (i.e., hexapod) learning state space.

\subsection{Influence of $\beta$}

An important parameter which needs to be discussed in our learning method is the annealing factor $\beta$.
It determines the learning time and whether it can converge globally.
In our learning process, if $\beta$ is set to 0 the acceptance probability is 1, i.e., every possible change of leg periods is acceptable. Therefore, the search is performed by random permutations.

If $\beta$ increases to positive infinity, the acceptance probability approaches zero. Therefore, the current combination of periods will be discarded only if the deviation is larger than the previous one, i.e., the learning algorithm does not accept worse solutions. In this condition, the simulated annealing method is reduced to a greedy search method.

\begin{figure}
	\begin{center}
		\includegraphics[height=2.4in]{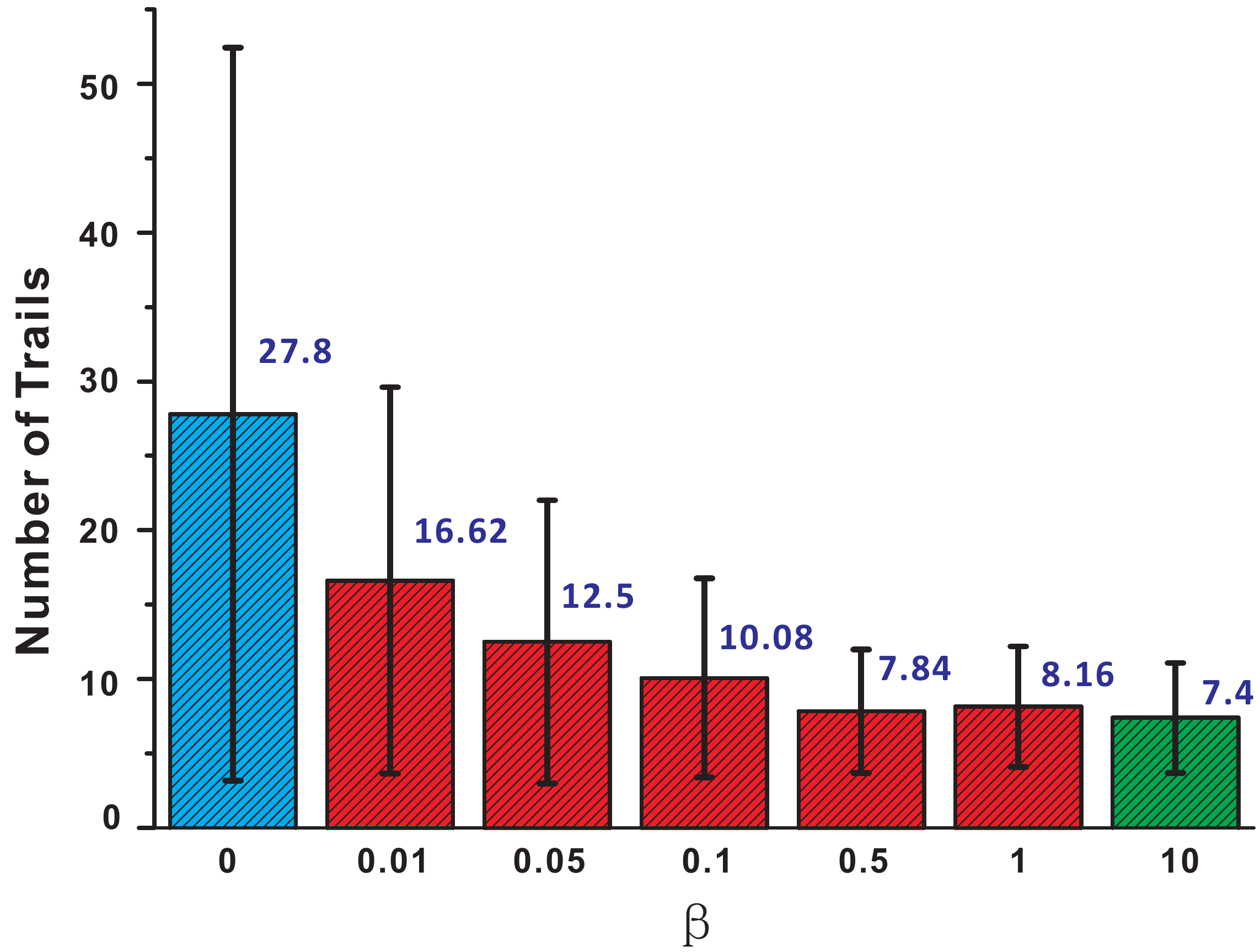}
	\caption{The average number of trials with different $\beta$ (see text for details). }
	\label{comp_of_beta}
	\end{center}
\end{figure}

Fig.~\ref{comp_of_beta} shows the number of trials when we set different annealing factors $\beta$. 
The blue bar represents random permutations ($\beta=0$) and the green bar represents approximately a greedy search ($\beta=10$). The red bars indicate results from the simulated annealing method when we use five different $\beta$. Every situation (i.e., every $\beta$) was tested for 50 times and the mean value is depicted by the blue numbers upon the bars. In this figure, the R1 and R2 leg were disabled while other legs were functional.

When $\beta=0$ (random permutation), we observed the longest search time. With increase of $\beta$, search time decreased, however, if $\beta$ was too large (larger than 10 here), learning might get stuck at a worse solution.
For example, in a scenario where R1 and L2  were disabled, the greedy search got stuck at a solution, where the remaining legs performed with periods R2=4, R3=4, L1=4 and L3=4. In this situation, the deviation angle remained at about $9.24^{\circ}$, which is larger than the predefined threshold of $8^{\circ}$.
No matter how we changed a leg's oscillation frequency, the deviation angle increased such that all combinations were discarded.
However, as shown in Fig.~\ref{learningresult}, the combination of R1=8, R3=4, L1=4 and L2=8 (line 11) would have been a suitable solution (found by simulated annealing, $\beta = 0.5$).

In summary, greedy search has a fast convergence speed but is more susceptible to getting stuck at a solution which might be worse than the solution obtained by SA. That is, the average deviation angle from the greedy search might be still larger than the one obtained from the simulated annealing. Employing random permutations will converge globally but costs more time. The simulated annealing technique lies between greedy search and random permutations. In principle, it is globally convergent and the search time is short if we employ an appropriate annealing factor \cite{granville1994simulated}. Finally, SA is able to find a suitable combination of periods for compensating leg malfunction. Therefore, the simulated annealing method is a good choice for our learning process.

\section{Real robot experiments}
\label{sect5}
\subsection{The walking machine platform AMOSII}

\begin{figure}[t]
\begin{center}
	\subfigure[Walking platform]{
      \label{fig1:a}
	\includegraphics[width=2.3in]{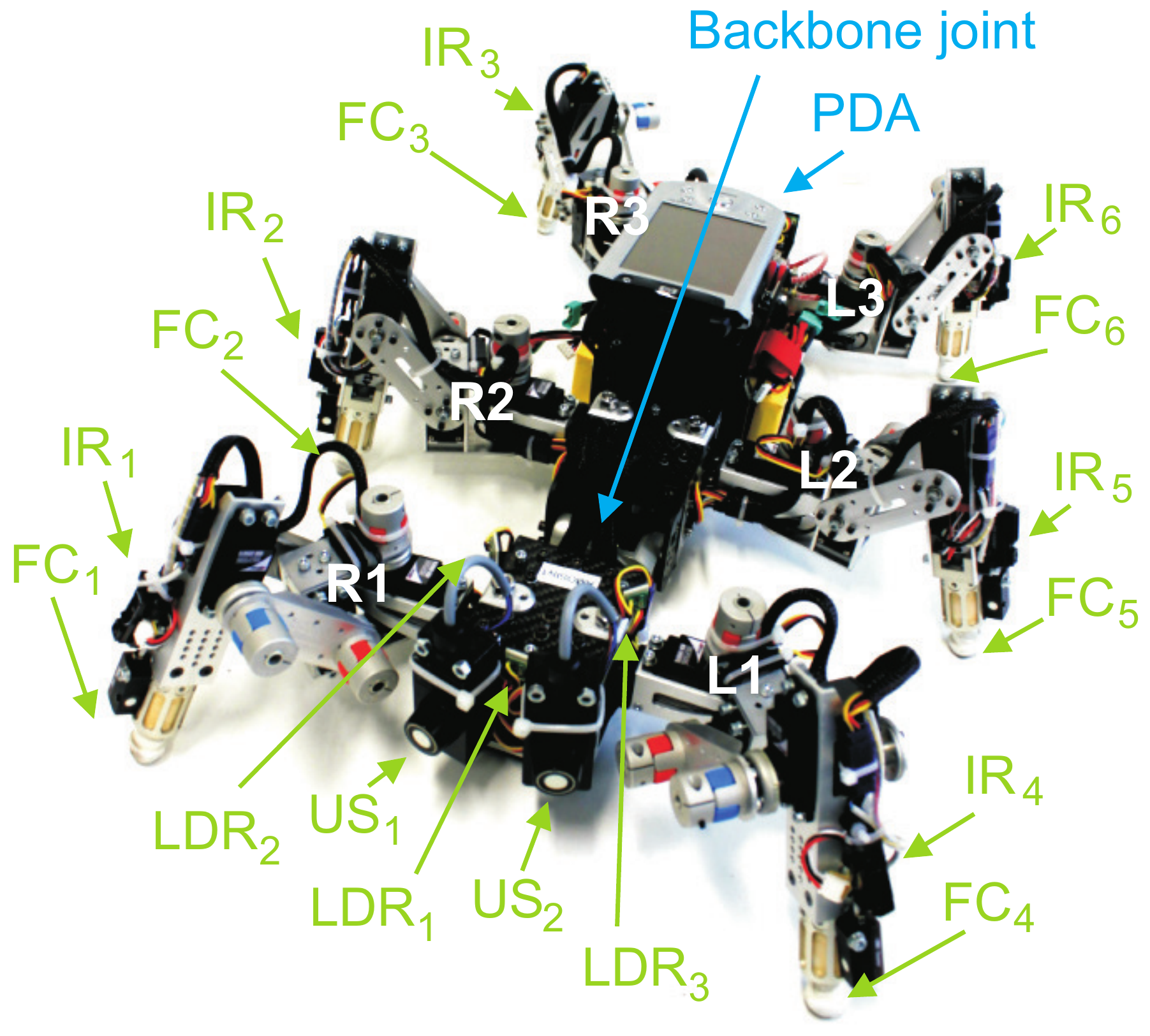}
	}
	\subfigure[Leg structure]{
      \label{fig1:b}
	\includegraphics[width=2.3in]{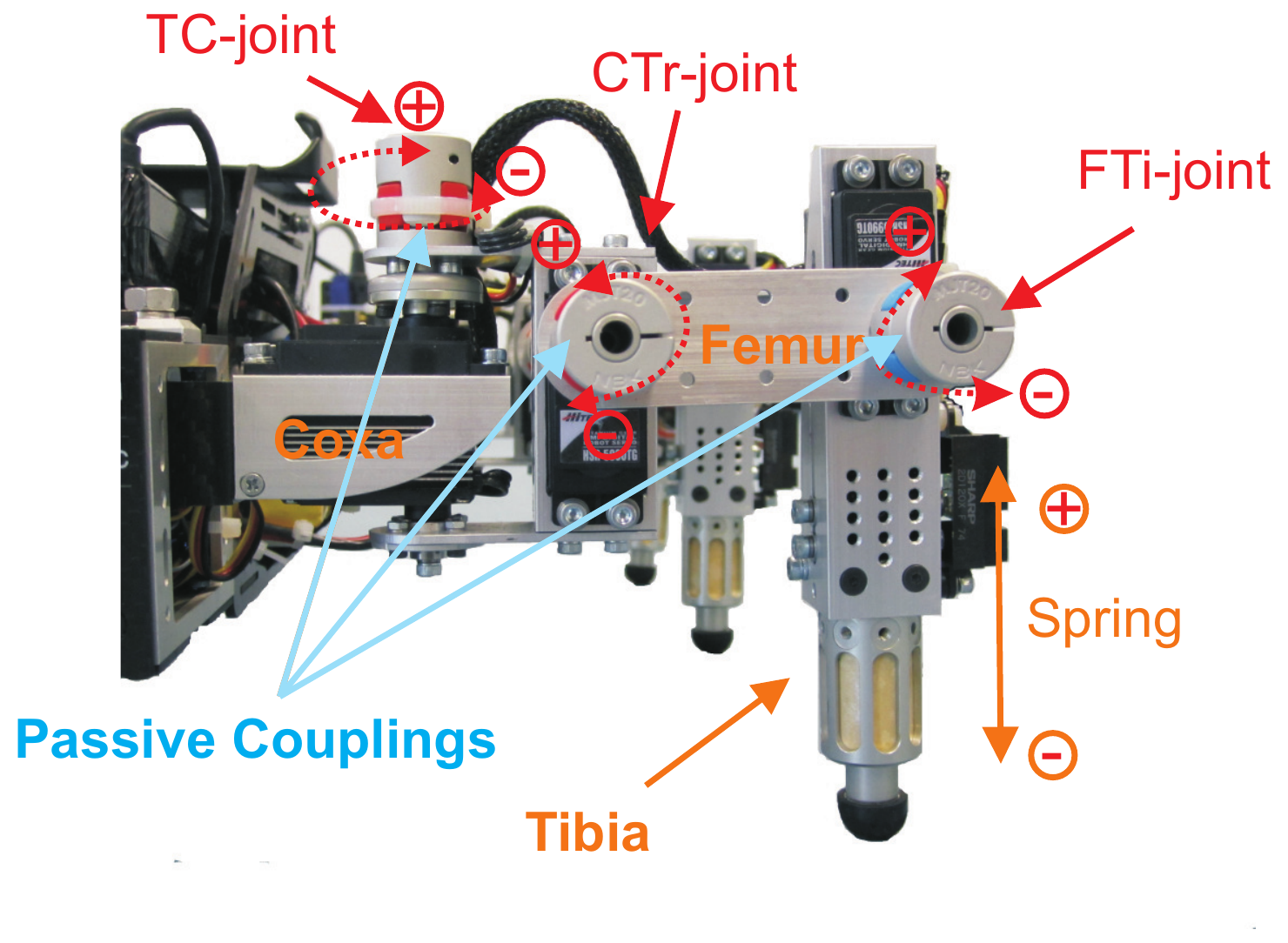}
	}
	\caption{
(a) Biologically inspired walking machine platform AMOS\uppercase\expandafter{\romannumeral2} with sensors. (b)
Leg structure inspired by a cockroach leg.}
	\label{fig1}
\end{center}
\end{figure}

In order to test our algorithm in a physical system, the six-legged walking machine AMOS\uppercase\expandafter{\romannumeral2}\footnote{AMOS\uppercase\expandafter{\romannumeral2} was developed by Bernstein Center for Computational Neuroscience at Georg-August-Universit\"at G\"ottingen in collaboration with Fraunhofer Institute IAIS, Germany.} is employed (see Fig.~\ref{fig1:a}).
It has identical leg structure with three linkages (coxa, femur, and tibia, see Fig. \ref{fig1:b}).
Each leg has three joints: the thoraco-coxal (TC-) joint enables forward ($+$) and backward ($-$) movements, the coxa-trochanteral (CTr-) joint enables elevation ($+$) and depression ($-$) of the leg, and the femur-tibia (FTi-) joint enables extension ($+$) and flexion ($-$) of tibia.
Compared to a real insect \cite{delcomyn1999walking}, the tarsus is ignored in the current design.
In general the tarsus is for absorbing outside impact forces and to stick the leg to a walking surface \cite{cruse2009principles}.
Nevertheless, a spring is installed in the leg to substitute part of the function of the tarsus; i.e., absorbing the impact force during touchdown on the ground.
In addition, a passive coupling is installed at each joint (see Fig. \ref{fig1:b}) in order to yield passive compliance and to protect the motor shaft. The body consists of two parts: two front legs on the front part, and the middle and hind legs on the hind part. The two body parts are connected by an active backbone joint which enables the rotation around the lateral or transverse axis. This backbone joint is mainly used for climbing which is not the main focus here (but see \citet{goldschmidt2012biologically}). All leg joints as well as the backbone joint are driven by digital servomotors.

The robot has six infrared sensors ($IR_{1,...,6}$) at its legs, six force sensors ($FC_{1,...,6}$) in its tibiae, three light dependent resistor sensors ($LDR_{1,2,3}$) arranged in a triangle shape on the front body part, and two ultrasonic sensors ($US_{1,2}$) at the front body part (see Fig. \ref{fig1:a}). The force sensors are for recording and analyzing the walking patterns. The infrared sensors are used for detecting obstacles near the legs and the ultrasonic sensors are used for detecting obstacles in front. The light dependent resistor sensors serve to generate positive tropism like phototaxis.
We use a Multi-Servo IO-Board (MBoard) installed inside the body to digitize all sensory input signals and generate a pulse-width-modulated signal to control servomotor position. The MBoard can be connected to a personal digital assistant (PDA) or a personal computer (PC) via an RS232 interface. For the robot walking experiments presented here, the MBoard is connected to a PC on which the neural controller is implemented. Electrical power supply is provided by batteries: one 11.1 V lithium polymer 2,200 mAh for all servomotors and two 7.4 V lithium polymers for the electronic board (MBoard) and for all sensors.

\subsection{Experimental results}

\begin{figure}
	\begin{center}
		\includegraphics[width=0.91\linewidth]{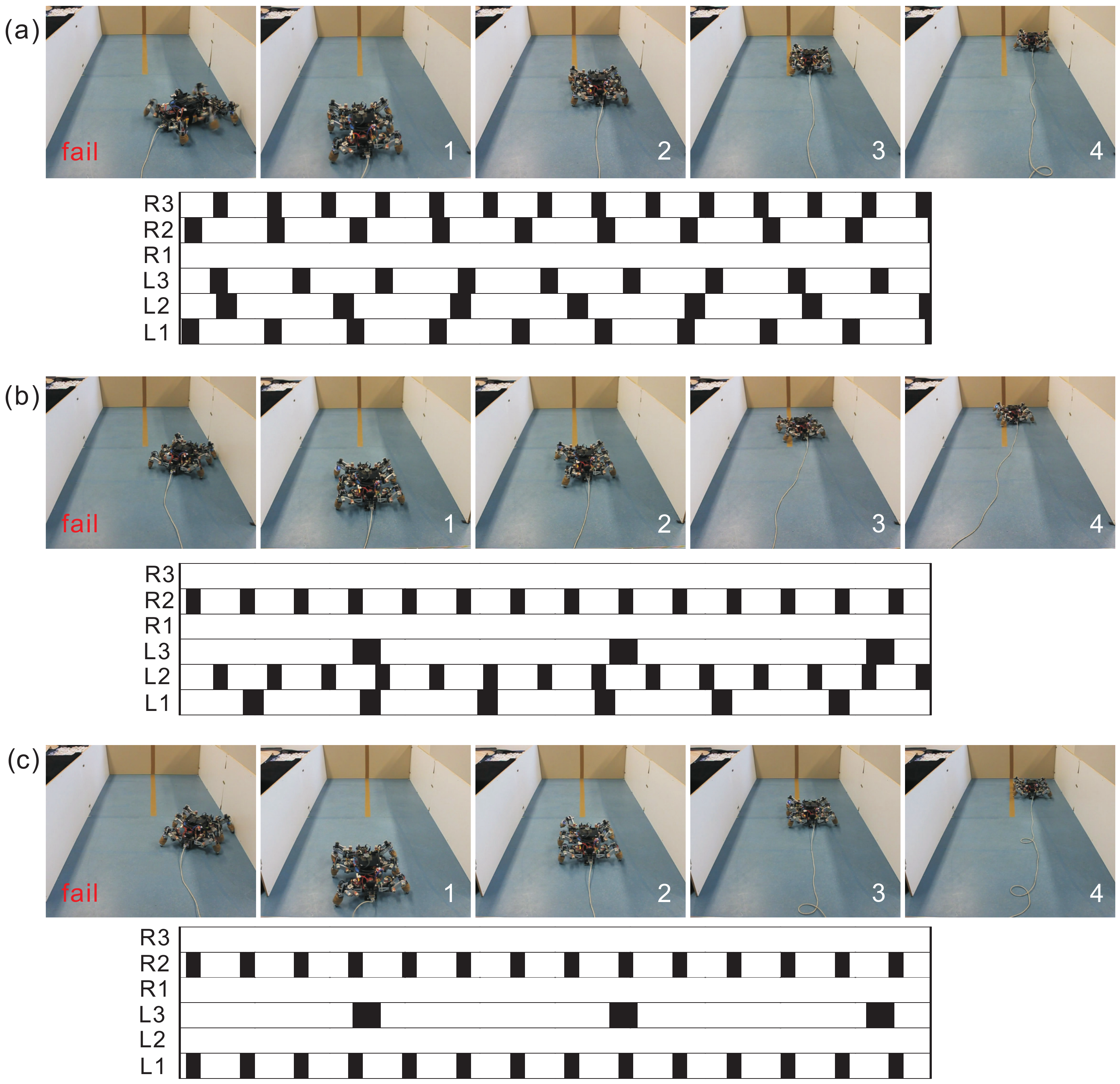}
	\caption{Three scenarios of the real robot experiments. For each subfigure, the upper
panel shows one snapshot of a fail situation (before learning) and four snapshots of a success
situation (after learning). The lower panel shows the gait (i.e., suitable leg frequencies) after learning. The gait is observed from the motor signals of the CTr-joints. A black area means that the leg touches the ground, while a white area indicates that the leg is in the air. (a) R1 leg disabled. (b) R1 and R3 legs disabled. (c) R1, R3, and L2 legs disabled. }
	\label{expresult}
	\end{center}
\end{figure}

The real robot experiments were conducted to test the validity of the proposed multiple CPGs and learning algorithm.
The controller is implemented with the update frequency of 27Hz.
The robot was placed in front of a tunnel which was 300 cm long and 120 cm wide (see Fig.~\ref{expresult}).
Three scenarios were tested: in the first, second, and third experiment, we disabled the R1 leg, the R1 and R3 legs, and the R1, R3, and L2 legs, respectively.
Using default periods (i.e., all legs moved with period 4), the robot deviated to the right and hit the right board in all three experiments, i.e., it failed to pass the tunnel.
For the R1 leg disabled, the robot deviated to the right by 48 cm after travelling 53 cm.
We reset the robot to the initial position and implemented the learned suitable combination of leg periods, i.e., 5, 4, 5, 6, 5 (obtained directly from the simulation, see line 1 in Fig.~\ref{learningresult}, in the sequence of R2, R3, L1, L2, L3).
These periods enabled the robot to pass the tunnel with a small deviation of approximately 42 cm.
For the R1 and R3 legs disabled using the default periods, the robot deviated 47 cm after travelling 120 cm.
We used the learned combination of 4, 6, 4, 9 (see line 8 in Fig.~\ref{learningresult}, in the sequence of R2, L1, L2, L3) and observed only a small deviation of 5 cm.
For the R1, R3, and L2 legs disabled using the default periods, the robot deviated 45 cm after travelling 105 cm.
In this situation, we used the learned periods 4, 4, 9 obtained directly from the simulation (see line 11 in Fig.~\ref{learningresult}, in sequence of R2, L1, L3), resulting in a deviation of 14 cm.
The deviations are also shows as angles in Table~\ref{tab1}.
Experimental snapshots and gait diagrams are shown in Fig.~\ref{expresult}. The experimental video can be seen in the supplementary material and at our website \url{http://manoonpong.com/MultiCPGs/supple_video.wmv}.
The video shows that the results from our simulations can directly transfer to the real robot resulting in the malfunction compensation in the robot's locomotion.


The experiments demonstrate the effectiveness of the proposed multiple CPGs with the learning mechanism. Additionally, as it is flexible, it can be easily extended to configurations other than the six CPGs implemented here; it can also be extended to 4 CPGs for the quadruped robot as shown in the simulation and 8 CPGs for a possible octopod robot in future research.

\begin{table}
\caption{The deviation angle (deg) before learning and after learning.}
\label{tab1}
\begin{center}
\begin{tabular}{c c c c }
\hline\hline
Scenarios & Disabled leg(s) & Before learning & After learning\\
\hline
1 & R1 & $\approx 42^{\circ}$ & $\approx 7.97^{\circ}$ \\
2 & R1, R3 & $\approx 21^{\circ}$ & $\approx 0.95^{\circ}$ \\
3 & R1, R3, L2 & $\approx 23^{\circ}$ & $\approx 2.67^{\circ}$ \\

\hline\hline
\end{tabular}
\end{center}
\end{table}

\section{Discussion}
\label{sect6}

Our contribution here is an extension of our previous works \cite{steingrube2010self,ren2012multiple,ren2013fault}. In \citet{steingrube2010self}, we proposed a single chaotic CPG for generating multiple gaits and complex behaviors, but excluded leg malfunction compensation. In \citet{ren2012multiple}, we investigated three chaotic CPGs with manual frequency tuning for malfunction compensation of a hexapod robot only. 
In \citet{ren2013fault}, only a short introduction to the work is given. 
The complete technical details, analysis, or experimental results presented here have not been published in our previous papers \cite{steingrube2010self,ren2012multiple,ren2013fault} or elsewhere.
In general, implementing such a CPG-based control strategy does not require the precise mechanical model. Furthermore, it does not need to calculate inverse kinematics,
which requires a lot computational resources.

The strategy of using multiple CPGs is inspired by the way insects control their locomotion. There were already experiments demonstrating that each insect leg is controlled by an independent oscillation center \cite{bassler1998pattern}. Additionally, they showed that legs are able to respond to their corresponding ganglions to generate rhythmic motion after amputation \cite{schilling2007hexapod,delcomyn1985factors}. Our control strategy is similar; the multiple CPGs are a decentralized system and based on a modular concept, with each leg controlled by an independent oscillator. Although our CPG model, exhibiting chaotic dynamics, and learning used here do not directly match biological findings, they form a powerful approach for robot locomotion control and learning. The multiple chaotic CPGs perform as open-loop controllers and only an orientation sensor is required and used as sensory feedback for learning. This results in a minimalistic system where learning can efficiently and automatically find a proper combination of oscillation frequencies of the different legs for malfunction compensation. Note that our approach is mainly developed to benefit technical systems like the walking robots used here. In real insects certain hard-wired sensori-motor loops are used instead of the learning mechanisms implemented in this study. For a robot however, the modeling of any such loop requires carefully addressing quite complex neuro-biological design issues \cite{fukuoka2003adaptive,ijspeert2007swimming}. Here, we demonstrate that learning can replace hard-wired sensorimotor loops, which for robotics substantially reduces design demands and we show that learning can indeed take the role of (evolutionary designed) sensori-motor structures regardless of the robot's specific embodiment (e.g., six- and four-legged robots shown here).

Some other approaches also successfully showed fault tolerance. For example, \citet{bongard2006resilient} presented an active process that allowed a robot to generate successful motor patterns for locomotion, before and after damage, through autonomous and continuous self-modeling. This algorithm was tested on a four legged starfish robot. \citet{schilling2007hexapod} improved the ``Walknet" by adding an analog-selector for dealing with leg amputations. An Artificial Immune System (AIS) was implemented on the OSCAR robot in order to detect leg anomaly and a Swarm Intelligence for Robot Reconfiguration (SIRR) method was applied to rearrange the body shape and to regulate the leg behaviors \cite{jakimovski2009self,el2006six,el2009maehle}.
\citet{spenneberg2004stability} realized fault tolerance locomotion of their octopod robot SCORPION by changing its gait to a hexapod gait.
\citet{yang1998fault} developed a fault tolerance mechanism based on a modeling and planning strategy to achieve stable walking with leg loss.
\citet{christensenfault} applied learning to automatically adjust a quadruped robot's gait such that fault tolerance and morphology optimization were realized. 
Compared to many of these approaches, we emphasize here a simple but robust CPG-based mechanism with learning for both multiple gait generation (see Fig.~\ref{gaitfig}) and leg malfunction compensation (see Fig.~\ref{learningresult}). The mechanism deals with only one parameter of each CPG for adaptation and requires only an orientation sensor as sensory feedback for learning.
Therefore, it converges faster compared to, e.g., the results shown in \cite{christensenfault}. 
We have also shown that it is easy to combine the learning mechanism with the multiple CPGs controller which is transferable to different platforms (see Fig.~\ref{learningresult} and \ref{fourleglearn}).

For our robot's locomotion control, especially for the malfunction compensation experiments, the center of mass of the body was maintained at a low level. In this configuration, the body sometimes touches the ground thus being able to support part of the body weight. Because of this, the remaining legs do not need to carry as much load and as a result the robot can move in a stable manner. This is similar to the way insects (e.g., cockroaches) perform their locomotion \cite{ritzmann2004convergent}. This configuration is especially effective when some of the robot's legs are disabled. In this situation, the functional legs are responsible for propelling the robot while the body is used to support the robot. If the robot meets with rough terrain, it can raise its body by depressing the legs to overcome the terrain irregularity. To compensate leg malfunction during walking on rough terrain, additional posture control might be required for balancing. However, investigating this issue is beyond the scope of this work.

When legs are disabled (one, two or even three), the resulting trajectory deviation can be compensated for by the proposed algorithm. The learning process starts as soon as the deviation exceeds the predefined threshold. Eventually the learning algorithm will find a suitable combination of periods. As the controller is modular, it is flexible and general \cite{hornby05autonomousevolution,valsalam:ieeetec11} and it can be extended from one CPG to the four or six CPGs presented here.
Furthermore, the modularity of the system offers the future possibility of combining it with other neural sensory processing modules to generate various sensor-driven behaviors (like obstacle avoidance, phototropism, not discussed here, see \cite{manoonpong2008sensor,steingrube2010self}). The modules are inhibited in case of leg malfunction, thereby not interfering with the learning process for leg malfunction compensation.

\section{Conclusion}
\label{sect7}

In this paper, we focus on the locomotion control of legged robots using neuron-inspired central pattern generators for the purpose of automatic malfunction compensation. This is a continuation of our previous investigations, where we had already demonstrated that a chaotic CPG could generate complex behaviors, such as different walking gaits as well as novel self-untrapping mechanism \cite{steingrube2010self,ren2012multiple}.

The main advantage of CPG based control is that it does not require a precise mathematical model of leg structure. However, with a single CPG, the robot cannot compensate for leg malfunction. As such, taking inspiration from multiple oscillators found in the real neural systems of insects, here we first extended our single chaotic CPG to multiple chaotic CPGs. The different CPGs can now synchronize to generate uniform frequency or desynchronize to oscillate independently. Here, robot leg malfunction causes the different CPGs to oscillate asynchronously. In this setup, in order to deal with the leg malfunction, an on-line learning mechanism based on the simulated annealing (SA) technique is applied. The SA algorithm allows the robot to automatically find a suitable combination of periods (frequency) for each leg and thereby considerably reduce any deviation from the original movement trajectory caused due leg malfunction. Furthermore, we show that such a SA based learning algorithm is capable of converging to a solution within acceptable time without getting stuck at a worse combination of periods. We clearly demonstrate the effectiveness and generalization of our learning algorithm using both simulations and real robot experiments.

In this study, although our control approach allows walking robots to learn to compensate for leg malfunction, individual joint malfunctions and body damage compensation have not been addressed. In the future, we will investigate these issues which might require more complex sensory information \cite{heliot2008multisensor,zhang2012effects} as well as additional mechanisms, such as advanced force control \cite{schilling2007hexapod,gorner2010analysis} and suitable muscle models \cite{xiong2013simplified} to enable leg compliance and adaptation \cite{der2001self}. Furthermore, cognitive abilities such as memory \cite{dasgupta2013information} and decision making \cite{dasgupta2013neural} will be also applied for goal-directed navigation learning. In addition, due to the employed chaotic CPGs controller which in principle can reliably obtain a stable (complex) periodic pattern out of a large (often infinite) number of unstable periodic patterns \cite{steingrube2010self}, we will also use the controller to deal with other motor tasks, like manipulation which requires a variety of complex patterns.

\section*{Acknowledgment}

This research was supported by the Emmy Noether Program (DFG, MA4464/3-1), the Federal Ministry of Education and Research (BMBF) by a grant to the Bernstein Center for Computational Neuroscience II G\"ottingen (01GQ1005A, project D1), the National Natural Science Foundation of China under the project 61175108, the Beijing Natural Science Foundation under the project 4142033, the Innovation Foundation of BUAA for PhD Graduates and the European
Communitys Seventh Framework Programme FP7/2007-2013 (Specific Programme Cooperation, Theme3, Information and
Communication Technologies) under grant agreement no.270273, Xperience.

We thank Dr. Frank Hesse and Xiaofeng Xiong for technical advice on simulator and real robot implementations, and we thank J\'er\'emie Papon for checking the language and proofreading of this paper.






\bibliographystyle{model1b-num-names}
\bibliography{ref}







\end{document}